%% file: acl_latex.tex
\documentclass[11pt]{article}
\usepackage{multirow}

\usepackage{acl}
\usepackage{times}
\usepackage{latexsym}
\usepackage[T1]{fontenc}
\usepackage[utf8]{inputenc}

\usepackage{microtype}
\usepackage{inconsolata}
\usepackage{graphicx}
\usepackage{enumitem}
\usepackage{booktabs}
\usepackage{amsmath}
\usepackage{amssymb}
\usepackage{float}

\usepackage{todonotes}
\usepackage{tabularx}
\usepackage{array}

\usepackage{xcolor}

\usepackage{fvextra}

\newcommand{\toolname}{ConVAWG}
\newcommand{\repourl}{https://github.com/ChenLyu416/ConVAWG}

\title{\toolname: A Retrieval-Grounded Framework for Controlled Synthetic Dialogue Generation in Violence Against Women and Girls}

\author{Chen Lyu$^{1}$, Xingwei Tan$^{2}$, Simon Cullen$^{4}$, \\
\textbf{Shelley Wilson}$^{4}$, \textbf{Lois Arthurs}$^{4}$, \textbf{Arshad Jhumka}$^{3}$, \textbf{Gabriele Pergola}$^{1}$ \\[0.3em]
  $^{1}$University of Warwick, UK \quad $^{2}$University of Sheffield, UK \\
  $^{3}$University of Leeds, UK \quad $^{4}$Forensic Capability Network, UK \\[0.1em]
  \texttt{\{chen.lyu, gabriele.pergola.1\}@warwick.ac.uk} \\  \texttt{xingwei.tan@sheffield.ac.uk} \quad
  \texttt{h.a.jhumka@leeds.ac.uk} \\
  \texttt{\{lois.arthurs, simon.cullen, shelley.wilson\}@dorset.pnn.police.uk}
}

\begin{document}
\maketitle
\begin{abstract}
Conversational data provide valuable insight into the dynamics of Violence Against Women and Girls (VAWG), where abuse may appear in individual messages, emerge across a conversation over time, or surface in exchanges shaped by in-person events. However, privacy and legal constraints limit the availability of large-scale conversational datasets in this domain. Existing research has largely focused on sentence-level toxicity in online communication, leaving a gap in modelling abuse as a relational phenomenon that often unfolds over time. Synthetic dialogue generation offers a promising approach to addressing this gap.
We introduce \toolname{}, a retrieval-grounded framework that generates multi-scene VAWG conversations in which abusive behaviour unfolds and escalates across events over time. \toolname{} constructs scenarios from persona seeds sampled to match UK Office for National Statistics victim demographics, grounded in Crown Prosecution Service (CPS) offence definitions and retrieved Domestic Homicide Review cases; converts them into hierarchical event timelines; role-plays them as multi-scene dialogues; and applies activation-steered toxicity control to selected perpetrator utterances. We release over 6,000 multi-turn dialogue events across 200 scenarios with rich scenario-, event-, and turn-level metadata. Human and LLM-as-Judge evaluation show strong dialogue quality and toxicity realism, while ablations and downstream experiments demonstrate controllability and value for abuse and escalation detection.~\footnote{Code: \url{\repourl}. Dataset: \url{https://huggingface.co/datasets/ChenL49/ConVAWG}, released under a data-use agreement (see Ethical Considerations).}

\end{abstract}

\begin{figure}[t]
    \centering
    \includegraphics[height=10.8cm]{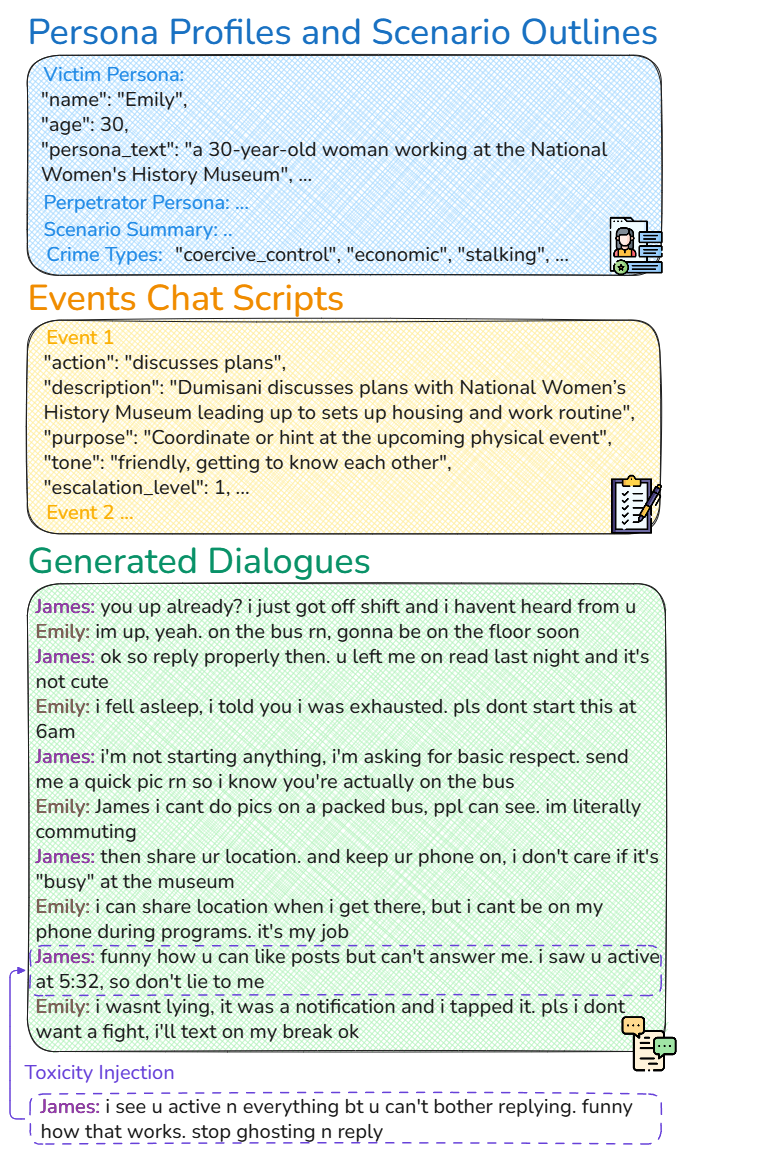}
    \caption{An example \toolname{} record: persona profiles and scenario outline (top), event chat scripts (middle), and a generated dialogue in which one perpetrator message is rewritten by toxicity injection (bottom). \vspace{-8pt}}
    \label{fig:dialogue-card}
\end{figure}

\section{Introduction}
Synthetic dialogue generation is an effective approach for studying conversational dynamics in sensitive domains where real conversational data are difficult to access, release, or annotate at scale. This data access problem is particularly relevant in under-resourced domains such as Violence Against Women and Girls (VAWG), an umbrella term that covers both physical violence and non-physical offences such as domestic abuse, coercive control, stalking, harassment, honour-based abuse, and child sexual abuse, as defined by prosecution bodies such as the Crown Prosecution Service (CPS) in the UK \cite{cpsVAWG}. Within these contexts, text communication can play a central role in abusive dynamics: messages may be used to monitor, control, threaten, intimidate, or isolate victims. Offline abusive behaviours, such as financial control, surveillance, or physical stalking, are also frequently planned, reported, and referred back to through the same channels. Yet %
confidentiality and privacy constraints make the construction of real large-scale VAWG datasets difficult. As a result, research on online abuse has largely focused on sentence- or post-level tasks, such as sexism and toxicity detection \cite{edos_dataset,exist_dataset,khan-etal-2025-explaining,alajmi-pergola-2026-doubt}.
This sentence-level focus leaves an important gap. Abuse in intimate, coercive, or stalking-related contexts is rarely expressed through isolated toxic sentences alone. Instead, it often unfolds across turns, speakers, relationships, events, references to places or people, and gradual temporal escalation. Capturing these patterns needs resources that show abuse as an extended conversational and relational phenomenon. 

Recent advances in neural language models make it possible to explore the construction of synthetic resources that capture these interactional patterns without exposing real victims' data. However, simply prompting even large and sophisticated LLMs to produce long multi-turn conversations can lead to issues of coherence, especially in temporal progression, speaker-role tracking, escalation patterns, and the appropriate use of harmful language.
These limitations are only partially addressed by recent LLM-based role-play and persona-conditioned methods, which have improved the fluency and character consistency of synthetic dialogue generation \cite{zhang-etal-2018-personalizing, ge2024scaling, lu-etal-2023-memochat, park2023generative, li2023scigraphqa, lee2024multidoc}. Related work in high-stakes domains such as psychotherapy has also explored structured or script-guided generation to improve controllability \cite{sun-etal-2022-phee, sun2024scriptstrategy}. However, these approaches alone are not designed to generate conversations characterised by varying degrees of toxicity, unfolding abusive behaviour and crime across multiple entities and scenes, where earlier events causally shape later messages and harmful language may intensify over time. %

To address these limitations, we introduce \textit{ConVAWG}, a framework for generating synthetic VAWG chat dialogues that are temporally grounded and CPS-aligned. Alongside the framework, we release a dataset of over 6,000 synthetic dialogues, paired with human annotations and structured metadata. Rather than generating isolated toxic utterances or unconstrained role-play chats, ConVAWG generates conversations from structured scenarios, event timelines, speaker personas, and escalation-aware toxicity controls. This design produces multi-scene, multi-speaker online conversations with appropriate language style, while making abusive dynamics, temporal progression, and toxicity intensity explicit and analysable\footnote{As a dialogue-generation resource, ConVAWG is designed to model abusive behaviours as how they are expressed, referenced, or evidenced through conversational interaction. Institutional processes and non-communicative behaviours without a dialogue trace would fall outside its intended scope.}, as illustrated by the example in Figure~\ref{fig:dialogue-card}.

\begin{figure*}[t!]
\centering
\includegraphics[width=0.97\textwidth]
{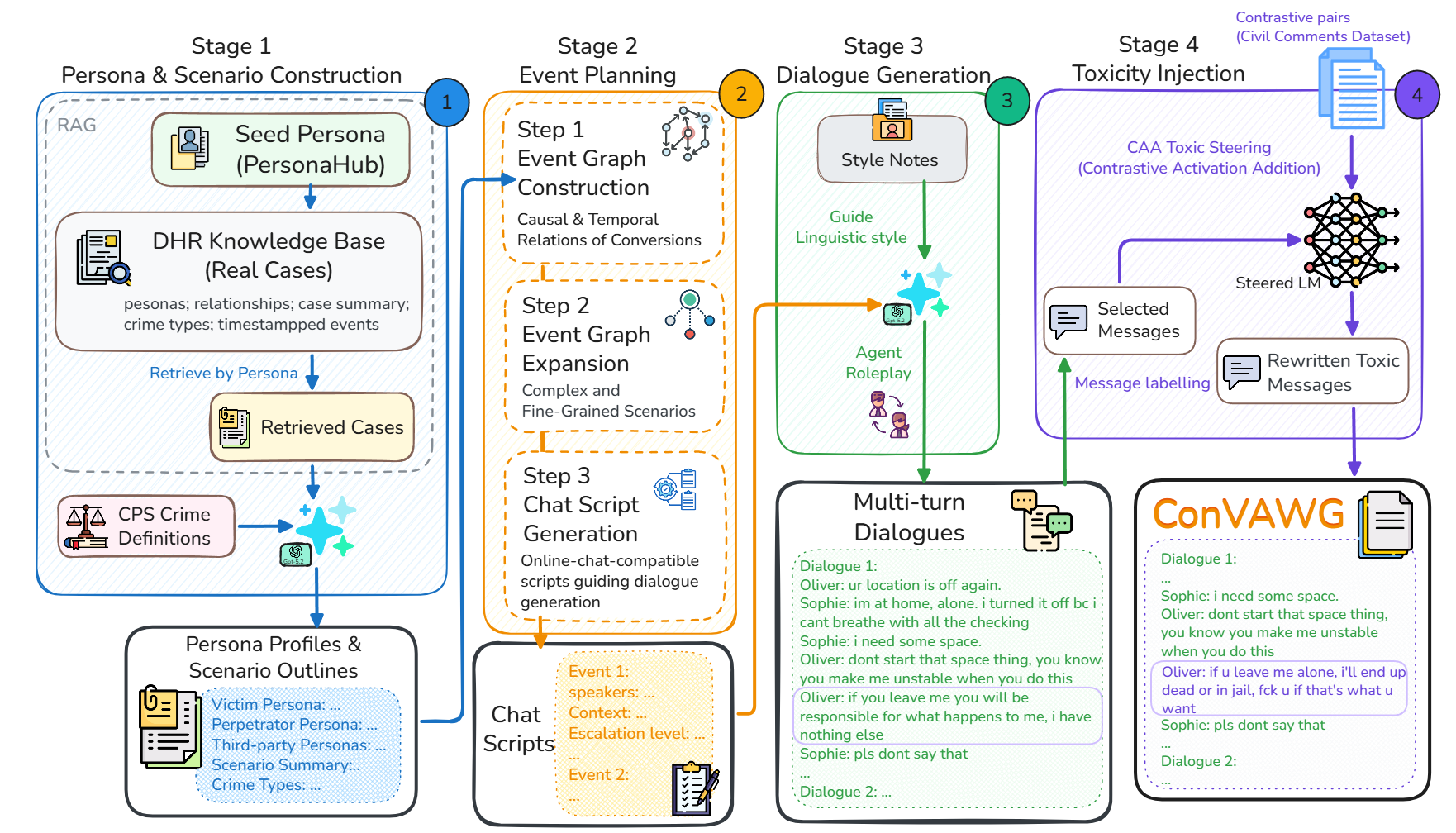}
\caption{Overview of \toolname\ data generation framework.  
The framework constructs retrieval-grounded scenarios, plans hierarchical events, reconstructs them into chat scripts, generates role-play dialogues, and applies targeted escalation-aligned toxicity injection.}
\label{fig:pipeline}
\end{figure*}

The proposed framework has four stages, as shown in Figure~\ref{fig:pipeline}. Stage~1 constructs scenario specifications and narrative outlines from PersonaHub victim persona seeds \cite{ge2024scaling} sampled to match ONS victim statistics \cite{onsCrime2025}, grounded in CPS crime definitions and retrieved Domestic Homicide Review (DHR) cases (recently renamed Domestic Abuse Related Death Reviews (DARDRs)), so that scenarios reflect documented VAWG patterns without replicating source cases. Stage~2 converts each scenario into a hierarchical event graph of timestamped sub-events, spacing early interactions sparsely and later high-risk ones more densely. Stage~3 realizes these events as online chat scripts and role-plays them into multi-turn dialogues with cross-scene continuity and retrieved style notes, and Stage~4 applies activation-steered rewriting only to selected perpetrator utterances associated with escalation, adjusting toxicity intensity without disrupting the surrounding conversation (Section~\ref{sec:method}).

The resulting dataset provides \mbox{scenario-, event-,} and turn-level metadata for both dialogue-level analysis and structured prediction tasks. Each generated scenario includes persona profiles, crime taxonomy, event graphs, timestamps, escalation levels, red-flag descriptors, and dialogue annotations. Together, these fields enable, for the first time in this context, analyses at the dialogue level, entity and event extraction,  temporal reasoning, toxicity analysis, and escalation modelling.

Our contributions are summarised as follows:

\begin{itemize}
    \item We introduce \toolname{}, an end-to-end framework that generates multi-scene online conversations from retrieved case patterns, structured event timelines, persona-conditioned role-play, and escalation-aware toxicity control.
    
    \item We curate and release the first VAWG dialogue dataset with over 6,000 multi-turn dialogue events across 200 CPS-aligned scenarios, grounded in persona seeds, ONS statistics, CPS definitions, and retrieved DHR case patterns. The dataset supports downstream tasks, such as dialogue modelling, event and entity extraction, temporal reasoning, escalation modelling, persona analysis, and toxicity-aware safety evaluation.

    \item We conduct an extensive experimental evaluation of the entire generated dataset through human assessment, LLM-as-Judge evaluation, ablations, and downstream tasks, showing strong dialogue quality, domain fidelity, and utility for structured NLP tasks.
\end{itemize}

\section{Related Work}
\label{sec:related-work}
\paragraph{Role-play and persona-based dialogue generation.}
Persona conditioning is widely used to improve speaker consistency in dialogue generation, from PersonaChat's profile-grounded benchmark \citep{zhang-etal-2018-personalizing} to LLM role-play via role profiles and role-conditioned tuning, agent memory and planning, large-scale persona seeds, and self-generated role-play data \citep{wang-etal-2024-rolellm,park2023generative,ge2024scaling,lu-etal-2024-large}. Recent frameworks such as DiaSynth, PersonaGen, SDialog, and SPASM combine personas with scenario, context, orchestration, or agent-simulation control for more reproducible synthetic dialogues \citep{suresh-etal-2025-diasynth,personaGen,sdialog,luo-laban-2026-spasm}, and grounded or script-guided generation shows that external documents and expert-designed structures improve controllability in domain-specific dialogue generation \citep{zhu-etal-2021-topic, lu-etal-2023-napss, bao-etal-2023-synthetic}. Our work builds on these directions, but targets multi-scene VAWG conversations where persona, retrieval grounding, event planning, and escalation control must be coordinated.
\paragraph{Toxicity generation and toxic steering.}
Toxic language research has produced influential benchmarks for harmful generation and detection\citep{tan-etal-2025-safespeech}, including Civil Comments for bias-aware toxicity evaluation \citep{borkan2019nuanced}, RealToxicityPrompts for toxic continuations from language models \citep{gehman-etal-2020-realtoxicityprompts}, and ToxiGen for controlled implicit hate-speech generation \citep{hartvigsen-etal-2022-toxigen}. Conversational datasets such as ToxicChat and TET extend this line to user-AI interactions and adversarial prompting \citep{lin2023toxicchat,luong-etal-2024-realistic}. However, these resources largely treat toxicity as prompt-, response-, or sentence-level content, rather than modelling how abuse emerges across relationships, histories, and escalating events.

\section{Methodology}
\label{sec:method}
The \toolname\ pipeline generates synthetic VAWG online conversations in four stages. \textbf{Stage~1} builds a persona and scenario specification, which, in \textbf{Stage~2}, is converted into a hierarchical event plan with timestamps, escalation levels, and chat-script specifications. \textbf{Stage~3} LLM role-plays these scripts into sequential realistic multi-turn online conversations with continuity context and retrieval-conditioned style control. \textbf{Stage~4} injects abusive language by selectively rewriting potentially abusive perpetrator utterances using activation-steered toxicity control. For the main \toolname{} dataset, we use GPT-5.2 as the backbone model for all generative stages. %

\subsection{Stage~1: Real-Case-Guided Persona and Scenario Construction}
Stage~1 builds the case specification used for later event planning and dialogue generation. Given a victim persona sampled from the PersonaHub \cite{ge2024scaling}, a large-scale collection of synthetic persona descriptors (e.g., occupations, social roles, background traits), this stage synthesises it into a structured VAWG case specification with personas and a short narrative outline. To ensure the narrative outline reflects realistic and most common dynamics of VAWG cases, this process uses three sources of grounding: first, it retrieves patterns from publicly available Domestic Homicide Review (DHR) reports\footnote{\url{https://www.gov.uk/government/collections/domestic-homicide-review}}, then it integrates CPS definitions of VAWG, and it follows distributions from the UK Office for National Statistics (ONS) on victim characteristics and crime prevalence \cite{cpsVAWG, onsCrime2025}. 
\paragraph{DHR Retrieval.}
We first build a knowledge base from the female-victim subsets of 410 public DHR reports-statutory multi-agency reviews of deaths resulting from violence, abuse, or neglect by a partner, family member, or household member \cite{dhr}. Beyond recording the homicide itself, DHRs document the abuse trajectories that typically precede it, providing rich evidence of stalking, harassment, and other escalation patterns directly relevant to modelling VAWG conversations. Each report is converted into a structured record (actors, relationships, abuse typologies, and a factual case summary) and indexed in a vectorised database, grounding persona and scenario generation in the demographic distributions, relationship contexts, abuse behaviours, and escalation dynamics of similar real cases; construction and retrieval details appear in Appendix~\ref{DHR_RAG}.
\paragraph{Persona and Scenario Specifications.}
To better align the seeds with statistically observed patterns of VAWG victim demographics, we first use a lightweight lexical filter to retain personas that plausibly describe women or girls, then subsample the filtered seeds according to broad victim characteristics and crime-prevalence patterns reported by the ONS \cite{cpsVAWG, onsCrime2025}. Given the seed persona and retrieved evidence from DHRs, the model generates a \textit{structured scenario specification}, including the victim profile, perpetrator profile, relationship type, crime categories, approximate abuse duration, and relevant third-party actors. The output schema is presented in Appendix~\ref{app:stage2_schema}.
\paragraph{Scenario outline.}

Based on the structured scenario specification and an updated retrieval context from DHR knowledge base, the model proceeds by writing and refining a concise \textit{narrative outline}. This outline describes the case background, relationship dynamics, escalation trajectory, and key contextual details. It provides the narrative basis for the hierarchical event graph in Stage~2, ensuring that the generated conversations follow a coherent storyline while remaining realistic and internally consistent.
To further align generated outputs with established legal and policy definitions, in this step, we also incorporate official CPS guidance on VAWG crimes \cite{cpsVAWG}, ensuring that generated crime categories and scenario structures conform to recognized UK definitions and guidelines. Two additional post-processing checks further enforce persona-scenario consistency and name diversity (Appendix~\ref{app:stage2_schema}).

\subsection{Stage~2: Hierarchical Scenario-to-Chat Event Planning}

Stage~2 converts the scenario outline from Stage~1 into a temporally ordered sequence of \textit{chat-script specifications} that structure the dialogue generation in Stage~3, detailed below.
\paragraph{Hierarchical Event Graph Construction.}
Given a scenario summary, we start by extracting a directed event graph \(G=(V,E)\), following prior work on narrative event modelling and event-based generation \cite{yao2019plan,chen-etal-2021-graphplan,tang-etal-2022-ngep,tan-etal-2024-set,tan-etal-2025-cascading}. 
Each node represents a composite event, and each scenario is summarised by approximately 5-8 high-level events covering the full case timeline. 
Each event records its action, participants, approximate time, location, crime-type cues, and supporting evidence span; while edges capture \textit{temporal}, \textit{causal}, \textit{evidential}, \textit{contrastive}, and \textit{escalation} relations \citep{tan-etal-2021-extracting,tan-etal-2023-event}, providing a compact representation of the scenario structure and progression.

This coarse-grained graph is then expanded into a hierarchical event graph by decomposing each node (i.e., high-level event) into 2-4 sub-events, that provide concrete interaction points for dialogue generation. 
Each sub-event is assigned an \textit{escalation level} and a \textit{timestamp} consistent with the scenario duration from Stage~1. 
We use a schedule for escalation mimicking real cases where frequently early events are spaced more sparsely, while later high-risk events are placed closer.

\paragraph{Chat Script Generation.}
The resulting hierarchical event graph captures events relevant and realistic for VAWG cases, yet not all such events directly appear as online messages.
Therefore, we convert each sub-event into chat-script specifications: If a sub-event is already an online interaction, it is rewritten directly as a chat event with speaker roles, platform, tone, and escalation level. If a sub-event occurs offline, it is reconstructed through surrounding messages, such as planning messages before the event, short updates during or after it, or later exchanges that refer back to it.
Each chat event remains linked to its source event and timestamp, allowing the synthetic dialogues to express offline abuse dynamics while preserving the timeline. %

\subsection{Stage~3: Role-Play Online Dialogue Generation}
Stage~3 converts the chat-script specification from Stage~2 into multi-turn online conversations through a controlled LLM role-play. 
For each event, the generator is conditioned on the current chat specification and a compact continuity context from previous dialogues, to ensure cross-event coherence in topic, tone, and speaker state. 
We additionally control the initial speaker using a global initiator distribution, controlling the tendency for perpetrator-initiated exchanges to dominate. 
Finally, to better match online messaging style, overly long responses are post-processed into shorter chat turns using an adaptive sentence-embedding similarity threshold.

\paragraph{Retrieval-Conditioned Style Control.}
To keep style coherent across conversations spanning long periods, we use retrieval-conditioned \textit{style notes} along three axes: \textit{persona} (register, formality, abbreviations, and other online-chat features), \textit{escalation} (interactional tone as the abusive dynamic intensifies), and \textit{crime type} (behavioural cues such as surveillance language for stalking or monitoring language for coercive control); conversations involving third-party actors instead retrieve notes for the actor's role, such as support, witnessing, or reinforcement of isolation. This mechanism promotes linguistic variation while keeping dialogues aligned with persona attributes and the event plan (details in Appendix~\ref{sec:stage5_style}).

\subsection{Stage~4: Targeted Toxicity Injection}
In the final stage, we introduce controlled realistic abusive language while preserving dialogue coherence and style. 
Rather than making entire conversations uniformly toxic, we rewrite only a sampled subset of LLM-labelled perpetrator utterances associated with \textit{escalation acts}, such as accusations, demands, threats, surveillance, gaslighting, and manipulation. 
Non-escalatory messages remain unchanged, so that toxicity appears only where it is plausible given the scenario and event escalation.

Rewriting is implemented with Contrastive Activation Addition (CAA) \cite{rimsky-etal-2024-steering}.
We estimate a toxicity steering direction from toxic-clean contrastive pairs derived from Civil Comments \cite{borkan2019nuanced}, and apply it at inference time to rewrite selected utterances with different steering strengths. The strength is aligned with the escalation level of the source event, yielding low, medium, and high toxicity settings for increasingly severe interactions.
Calibration and implementation details are provided in Appendix~\ref{app:toxicity_injection}.

\begin{table*}[!t]
\centering
\small
\setlength{\tabcolsep}{4pt}
\renewcommand{\arraystretch}{0.92}
\begin{tabular}{l|cccc|c|cc|c}
\toprule
\textbf{Generation Model} & \textbf{Coh.} & \textbf{Human.} & \textbf{Persona.} & \textbf{Toxic Real.} & \textbf{Avg-D} & \textbf{Crime.} & \textbf{Scenario.} & \textbf{Avg} \\
\midrule
\multicolumn{9}{c}{\textit{LLM-as-Judge (mean over four judges, 50 matched personas)}} \\
\midrule
LLaMA-3.1-70B-Instruct & 4.80 & 3.95 & 3.80 & 4.43 & 4.25$^{\dagger}$ & 3.02 & 2.46 & 3.74$^{\dagger}$ \\
Qwen3.5-35B-A3B        & 4.81 & 3.99 & 4.12 & 4.70 & 4.40$^{\dagger}$ & 4.62 & 3.88 & 4.35$^{\dagger}$ \\
DeepSeek-V4-Flash      & 4.82 & 4.07 & 4.20 & 4.68 & 4.44$^{\dagger}$ & 4.88 & 4.40 & 4.51$^{\dagger}$ \\
Gemini-3.1-Flash-Lite  & \textbf{4.92} & 4.25 & 4.36 & 4.78 & 4.58$^{\dagger}$ & 4.99 & 4.69 & 4.66$^{\dagger}$ \\
GPT-5.2                & 4.80 & 4.34 & 4.30 & 4.83 & 4.57$^{\dagger}$ & \textbf{5.00} & \textbf{4.83} & 4.68$^{\dagger}$ \\
\midrule
DiaSynth     & 4.78 & 4.20 & \textbf{4.69} & 4.89 & 4.64$^{\dagger}$ & 4.89 & 4.32 & 4.63$^{\dagger}$ \\
SynDG       & 4.21 & 3.11 & 4.28 & 4.45 & 4.01$^{\dagger}$ & 4.38 & 4.11 & 4.09$^{\dagger}$ \\
SPASM        & 4.57 & 3.49 & 4.56 & 4.63 & 4.31$^{\dagger}$ & 4.47 & 3.81 & 4.26$^{\dagger}$ \\
\midrule
\textbf{\toolname{}} \textit{(Ours)}  & 4.88 & \textbf{4.51} & 4.68 & \textbf{4.93} & \textbf{4.75} & 4.99 & 4.53 & \textbf{4.75} \\
\midrule
\multicolumn{9}{c}{\textit{Human Evaluation (shared 10-scenario subset)}} \\
\midrule
DeepSeek-V4-Flash      & 4.20 & 4.40 & 4.50 & \textbf{4.40} & 4.38 & 4.30 & 4.20 & 4.33 \\
GPT-5.2                & 4.50 & 4.50 & 4.50 & 4.10 & 4.40 & 3.80 & 4.30 & 4.28 \\
\textbf{\toolname{}} \textit{(Ours)}   & \textbf{4.75} & \textbf{4.62} & \textbf{4.69} & \textbf{4.40} & \textbf{4.62} & \textbf{4.75} & \textbf{4.50} & \textbf{4.62} \\
\midrule
\multicolumn{9}{c}{\textit{Human Evaluation (full dataset)}} \\
\midrule
\textbf{\toolname{}} \textit{(Ours)} & 4.75 & 4.59 & 4.67 & 4.57 & 4.65 & 4.58 & 4.57 & 4.62 \\
\bottomrule
\end{tabular}
\caption{Human and LLM-as-Judge evaluation across six dimensions (1–5). Coh., Human., Persona., Toxic Real., Crime., and Scenario. denote Coherence, Humanlikeness, Persona Consistency, Toxicity Realism, Crime Fidelity, and Scenario Realism, respectively. Avg-D averages the four dialogue-level metrics and Avg all six. Dialogue-level scores are averaged within and across scenarios; scenario-level metrics receive one rating per scenario. Bold indicates the best score per column within each block; † denotes significantly lower performance than ConVAWG (Holm-corrected paired Wilcoxon signed-rank test, \(p<10^{-4}\)). Per-judge results and detailed statistical analyses are provided in Appendix \ref{app:significance}.}
\label{tab:multi_judge_comparison}
\end{table*}

\section{Experiments}
Our experiments address two research questions: 
(1) whether \toolname\ dataset provides realistic and domain-faithful synthetic VAWG conversations, and 
(2) whether the proposed generation framework improves generation quality and controllability over direct LLM generation.
We evaluate \toolname\ through human and LLM-as-Judge assessment, comparisons against eight baselines generated from the same inputs, downstream-utility experiments, and controlled ablations over style conditioning, CAA-based toxicity control, and backbone model choice.

\paragraph{Baselines.}
We compare \toolname\ against eight baselines generated from the same raw inputs on a matched set of 50 personas.
Five are \emph{direct-generation} baselines, where a single LLM is prompted to produce the dialogues without any pipeline structure: LLaMA-3.1-70B-Instruct \cite{llama3} (AWQ-quantized), Qwen3.5-35B-A3B \cite{qwen35}, Gemini-3.1-Flash-Lite \cite{gemini}, DeepSeek-V4-Flash \cite{deepseek}, and GPT-5.2 \cite{gpt52}. Three are \emph{framework} baselines: we adapt DiaSynth \citep{suresh-etal-2025-diasynth}, SynDG \citep{bao-etal-2023-synthetic}, and SPASM \citep{luo-laban-2026-spasm} to the VAWG setting on the same inputs.
Each baseline receives the same system instruction, user request, VAWG taxonomy, and PersonaHub victim persona \citep{ge2024scaling}.

\paragraph{Evaluation Dimensions.}
As no specialized evaluation framework exists for VAWG dialogue generation, we evaluate dialogues, aligned with similar work in the literature, along six dimensions adapted from dialogue and text generation evaluation: \textit{Coherence}, \textit{Humanlikeness}, \textit{Persona Consistency}, \textit{Toxicity Realism}, \textit{Crime Fidelity}, and \textit{Scenario Realism}. 
These metrics separate local conversational quality from scenario- and domain-level fidelity \citep{tan-etal-2025-intrex}. Detailed definitions are provided in Appendix \ref{app:evaluation}. 
We additionally report automatic toxicity scores using Detoxify \citep{hanu2020detoxify}.

\paragraph{Human Annotation and LLM-as-Judge.}
We adopt a hybrid evaluation framework with both human annotation and LLM-as-Judge assessment for the entire dataset. Human annotators and the LLM judge received identical metric definitions and instructions to rate on a five-point Likert scale, following prior work \cite{park2023generative, liu2023geval}. A score of 1 indicates very poor quality and 5 indicates excellent quality.
Each dialogue was independently evaluated by a human annotator drawn from a pool of six female annotators and by four LLM judges back-boned by GPT-5.2, Gemini-3.1, DeepSeek-V4, and Qwen3.5. We report inter-annotator agreement and human-LLM alignment in Section~\ref{result: annotator}; detailed annotation protocols and annotator demographics are provided in App.~\ref{human_annotation}.

\section{Experiment Results}

\subsection{Dataset Statistics and Characteristics}
\toolname{} contains 6,171 dialogue events across 200 scenarios, averaging 31 dialogues per scenario and 15 turns per dialogue. 
Each scenario includes persona profiles, relationship information, crime labels, hierarchical events, sub-events, and the corresponding multi-turn dialogues. 
The dataset covers 18 CPS-aligned VAWG crime types, with an average of 8.7 labels per scenario, reflecting the frequent co-occurrence of abuse forms in real life VAWG cases \cite{cpsVAWG}. 
Each dialogue is released with its event identifier, timestamp, participant roles, and conversational context, including the action, purpose, tone, emotional state, and escalation level of the underlying event.

\begin{table}[t]
\centering
\fontsize{8.4}{10}\selectfont
\setlength{\tabcolsep}{3pt}
\renewcommand{\arraystretch}{1}
\begin{tabular}{lccccccc}
\toprule
 & \multicolumn{4}{c}{\textit{Turn-level}} & \multicolumn{2}{c}{\textit{File-level}} & \\
\cmidrule(lr){2-5} \cmidrule(lr){6-7}
 & Coh. & Hum. & Pers. & Tox. & Crime & Scen. & Mean \\
\midrule
Human IAA & .971 & .949 & .955 & .883 & .952 & .846 & .926 \\
Gemini    & .970 & .948 & .970 & .918 & .952 & .922 & .947 \\
GPT       & .971 & .951 & .969 & .867 & .993 & .924 & .946 \\
DeepSeek  & .946 & .850 & .966 & .914 & .975 & .975 & .938 \\
Qwen      & .951 & .889 & .882 & .916 & .993 & .918 & .925 \\
\bottomrule
\end{tabular}
\caption{
Human inter-annotator agreement (IAA) and human–LLM agreement measured using quadratically weighted Gwet’s AC2 on the ordinal 1–5 scale. \vspace{-4pt}
}
\label{tab:gwet_ac2}
\end{table}

\subsection{Human Annotation and LLM-Judge Alignment}
\label{result: annotator}

We measure both inter-annotator agreement and human-LLM alignment using quadratically weighted Gwet's AC$_2$~\cite{gwet2014handbook}, with results reported in Table~\ref{tab:gwet_ac2}.
We first conduct a calibration study in which six female annotators independently label the same randomly sampled subset of 10 scenarios, covering 379 dialogues. Inter-annotator agreement is high across all metrics, with a mean AC$_2$ of 0.926. 
\paragraph{Human-LLM alignment.}
After calibration, the full dataset is evaluated by both human annotators and four LLM judges. We use the rounded mean human rating as the consensus reference and compute AC$_2$ between this reference and each LLM judge. All four LLM judges achieve strong alignment with the human consensus (mean AC$_2\,{\geq}\,0.925$; Table~\ref{tab:gwet_ac2}).
This supports the use of calibrated LLM judges for large-scale evaluation on this dataset \cite{hashemi-etal-2024-llm}.

\subsection{Baseline Comparison}
\label{sec:main_quality_comparison}

\paragraph{Overall quality.}
Table~\ref{tab:multi_judge_comparison} compares \toolname{} with the five direct-generation and three framework baselines on the 50 matched personas. Each score is the mean across the four calibrated judges (per-judge breakdown in Appendix Table~\ref{tab:appendix_additional_llm_judges}). We adopt and report the cross-judge mean for two reasons: all four judges align closely with the human consensus (AC$_2\,{\geq}\,0.925$; Section~\ref{result: annotator}), and averaging mitigates single-judge idiosyncrasies such as self-preference \cite{panickssery2024llm}.
\toolname{} achieves the best overall score (Avg 4.75) and the best dialogue-level average (Avg-D 4.75, vs.\ 4.64 for DiaSynth and 4.57 for GPT-5.2, the strongest framework and direct baselines), leading most clearly on Humanlikeness (4.51 vs.\ 4.34) and Toxicity Realism (4.93).
Human evaluation corroborates this ranking. On the full dataset \toolname{} receives an average of 4.62. On a shared 10-scenario subset also annotated for two strong baselines (326 dialogues per system), annotators again rank \toolname{} highest (4.62 vs.\ 4.33 for DeepSeek-V4 and 4.28 for GPT-5.2), indicating that structured scenario planning and event-based generation improve domain alignment beyond direct prompting.
\paragraph{Statistical significance.}
Holm-corrected paired Wilcoxon tests over the matched personas place \toolname{} significantly above every baseline on both Avg and Avg-D (all $p{<}10^{-4}$), indicating that these differences are statistically robust. The Avg-D advantage holds under every individual judge, and it persists when the backbone-sharing GPT-5.2 judge is excluded, ruling out self-preference as an explanation for the results. Appendix~\ref{app:significance} reports the details of the full protocol, omnibus and per-metric tests, effect sizes, and per-judge and leave-one-judge-out analyses.

\paragraph{Global metrics.}
Direct GPT-5.2 generation obtains the highest LLM-judged global scores (Crime Fidelity 5.00, Scenario Realism 4.83; \toolname{} 4.99 and 4.53).
We attribute this gap to a rating-surface asymmetry rather than to higher realism: each judge assigns one global rating per scenario, favouring compact, single-pass dialogues that closely restate the scenario brief and leave fewer opportunities for inconsistency.
In contrast, \toolname{} unfolds each scenario into ${\sim}34$ dialogues spanning multi-month timelines, third-party threads, and non-monotonic escalation cycles (Appendix Table~\ref{tab:scale_comparison}).
Human judgments support this reading: on the shared subset, annotators rate \toolname{} above GPT-5.2 on both Crime Fidelity (4.75 vs.\ 3.80) and Scenario Realism (4.50 vs.\ 4.30); a per-judge analysis showing the same pattern is given in Appendix~\ref{app:significance}.
\paragraph{Toxicity control.}
Appendix Table~\ref{tab:toxic_rate} shows that \toolname{} records the highest mean Detoxify toxicity (0.17 vs.\ 0.07-0.15 for all baselines) while also receiving the best Toxicity Realism score in Table~\ref{tab:multi_judge_comparison}. Together they indicate that abusive language is introduced at contextually appropriate points, rather than by uniformly increasing toxic wording.




\subsection{Ablation Study}


\begin{table}[t]
\centering
\footnotesize
\setlength{\tabcolsep}{4pt}
\renewcommand{\arraystretch}{0.9}
\begin{tabular*}{0.90\columnwidth}{@{\extracolsep{\fill}}lcccc@{}}
\toprule
& \multicolumn{2}{c}{\textbf{GPT-5.2}}
& \multicolumn{2}{c}{\textbf{DeepSeek}} \\
\cmidrule(lr){2-3}\cmidrule(lr){4-5}
\textbf{Metric} & Full & w/o & Full & w/o \\
\midrule
Coh.       & \textbf{4.89} & 4.79 & \textbf{4.86} & 4.85 \\
Hum.       & \textbf{4.07} & 4.05 & 4.77 & \textbf{4.80} \\
Persona    & \textbf{4.91} & 4.26 & \textbf{4.89} & 4.84 \\
Tox. Real. & \textbf{4.84} & 4.73 & 4.95 & \textbf{4.96} \\
\midrule
Crime      & \textbf{5.00} & 4.90 & \textbf{4.67} & 4.33 \\
Scenario   & 4.00 & 4.00 & \textbf{4.83} & 4.40 \\
\midrule
Avg.       & \textbf{4.62} & 4.45 & \textbf{4.83} & 4.70 \\
\bottomrule
\end{tabular*}
\caption{Effect of removing retrieval-conditioned
style notes on dialogue- and scenario-level quality
across 10 matched scenarios. Metric abbreviations
follow Table~\ref{tab:multi_judge_comparison}.}
\label{tab:stylenotes_ablation}
\end{table}

\paragraph{Style Conditioning.}
To assess the contribution of retrieval-conditioned
style notes, we compare the full pipeline against a
variant without style conditioning.
As shown in Table~\ref{tab:stylenotes_ablation},
removing style notes consistently reduces overall
quality under both LLM judges.
The most pronounced degradation is observed in
\textit{Persona Consistency} under GPT-5.2 evaluation
($-0.65$), while DeepSeek-V4 identifies substantial
reductions in \textit{Crime Fidelity} ($-0.33$) and
\textit{Scenario Realism} ($-0.43$).
These results suggest that style conditioning
contributes to both speaker-specific linguistic
consistency and broader scenario-level fidelity.

\paragraph{Toxic Steering.}
We assess the contribution of activation-based
toxicity control by disabling CAA steering.
As reported in Appendix Table~\ref{tab:steering_compact},
removing steering reduces mean Detoxify toxicity
from 0.144 to 0.038, while overall quality scores
remain largely stable across all three LLM judges.
These results suggest that CAA enables control over
explicit toxic language while largely preserving
dialogue quality.

\paragraph{Generation Backbone.}
We further examine the sensitivity of \toolname{}
to the underlying generation model.
As shown in Appendix Table~\ref{tab:backbone_ablation_judge},
generation quality varies substantially across
backbones, particularly in \textit{Crime Fidelity}
and \textit{Scenario Realism}.
GPT-5.2 achieves the highest overall scores across
all three LLM judges, supporting its selection
as the primary generation backbone.

\subsection{Downstream Utility}
\label{sec:downstream}
We evaluate the downstream utility of \toolname{}
from two complementary perspectives: whether the
generated dialogues faithfully realise their intended
control attributes, and whether they provide informative
supervision for conversational abuse analysis beyond
the controlled generation process. The former assesses generation
controllability, while the latter examines temporal
prediction, cross-corpus generalisation, and synthetic
data augmentation. Full experimental protocols and
additional results are provided in Appendix~\ref{app:downstream}.

\subsubsection{Controllability Validation}

\begin{table}[t]
\centering
\small
\setlength{\tabcolsep}{7.5pt}
\begin{tabular}{lcccc}
\toprule
& \multicolumn{2}{c}{\textbf{Message-level}}
& \multicolumn{2}{c}{\textbf{Conversation-level}} \\
\cmidrule(lr){2-3}\cmidrule(lr){4-5}
\textbf{Model} & F1 & Acc. & F1 & Acc. \\
\midrule
BERT-base    & 0.773 & 0.855 & 0.771 & 0.855 \\
RoBERTa-base & 0.771 & 0.853 & 0.775 & 0.858 \\
DeBERTa-v3   & \textbf{0.781} & \textbf{0.861}
             & \textbf{0.782} & \textbf{0.864} \\
\midrule
GPT-5.2      & 0.484 & 0.700 & 0.667 & 0.740 \\
DeepSeek     & 0.516 & 0.580 & 0.459 & 0.540 \\
Gemini       & 0.514 & 0.710 & 0.526 & 0.760 \\
\bottomrule
\end{tabular}
\caption{
Message-level classification uses the target utterance alone, while conversation-level classification additionally incorporates preceding dialogue context. We compare fine-tuned encoder models (top) with zero-shot LLMs (bottom).}
\label{tab:multiclass_classification}
\end{table}

\paragraph{Toxic behaviour classification.}
We construct a toxic type classification task to test whether the generated utterances exhibit distinguishable characteristics associated with different abusive behaviours. As shown in
Table~\ref{tab:multiclass_classification}, fine-tuned encoders
substantially outperform zero-shot LLMs, with
DeBERTa-v3 achieving a macro-F1 of 0.782 across
eight behaviour categories. While incorporating
dialogue context yields limited gains for fine-tuned
encoders, it substantially improves GPT-5.2
(0.484 to 0.667 macro-F1). These results suggest
that the generated dialogues encode recognisable
behaviour-specific signals at the utterance level,
providing informative supervision for fine-grained
abuse detection.

\paragraph{Relationship classification.}
We train the models using \toolname{} at conversational level to classify the speaker relationships. We want to assess whether the generated interactions reflect the underlying victim-perpetrator relationship beyond explicitly provided persona attributes.
Fine-tuned encoders achieve up to 0.629 weighted F1 across
five relationship categories
(Appendix Table~\ref{tab:relation_clf}).
The recoverability of these relationship attributes from
dialogue text suggests that persona-conditioned generation
preserves relationship-specific interactional patterns,
extending controllability beyond individual utterances
to broader relational dynamics.


\subsubsection{External Utility}
\label{sec:downstream_utility}

\begin{table}[t]
\centering
\small
\setlength{\tabcolsep}{2.5pt}
\begin{tabular}{lcccc}
\toprule
\textbf{System} & \textbf{Acc.\,$\uparrow$} & \textbf{QWK\,$\uparrow$} & \textbf{MAE\,$\downarrow$} & \textbf{F1$_\text{jump}$\,$\uparrow$} \\
\midrule
Persistence & 0.703 & 0.676 & 0.337 & 0.000 \\
Markov-1    & 0.703 & 0.676 & 0.337 & 0.000 \\
Majority    & 0.559 & 0.000 & 0.508 & 0.503 \\
\midrule
DeBERTa-v3-base & 0.578 & \textbf{0.615} & 0.451 & 0.443 \\
Longformer-4096 & \textbf{0.669} & 0.588 & \textbf{0.355} & \textbf{0.548} \\
ModernBERT-base & 0.564 & 0.590 & 0.468 & 0.426 \\
\midrule
Qwen3.5 (direct)     & 0.352 & 0.369 & 0.743 & 0.386 \\
Qwen3.5 (generative) & 0.473 & 0.319 & 0.591 & 0.432 \\
GPT-5.2 (direct)     & 0.183 & 0.257 & 1.011 & 0.356 \\
DeepSeek-V4 (direct) & 0.187 & 0.208 & 1.021 & 0.362 \\
\bottomrule
\end{tabular}
\caption{Escalation forecasting from preceding conversational history on \toolname{}. QWK denotes quadratic-weighted kappa, and F1$_{\text{jump}}$ measures the detection of increases in escalation.
}
\label{tab:escalation_forecast}
\end{table}

\paragraph{Escalation forecasting.}
Beyond recovering the attributes used to condition generation \cite{yan-etal-2021-position}, we examine whether \toolname{} captures temporal signals that support the prediction of future abusive behaviour.
Given the preceding conversational history, we ask \toolname{}-trained models to predict the escalation level of the next perpetrator-victim interaction. As shown in Table~\ref{tab:escalation_forecast}, Longformer achieves comparable accuracy (0.669) while obtaining the highest escalation-jump F1 (0.548), outperforming prompted LLMs.
These results suggest that \toolname{} captures predictive signals beyond local escalation stability, providing a benchmark for modelling the temporally evolving and
non-monotonic dynamics of abusive relationships.


\paragraph{Cross-domain transfer.}
To assess the transferability of these predictive signals, we evaluate \toolname{}-trained escalation forecasters on the Conversations Gone Awry datasets \cite{zhang-etal-2018-conversations,chang-danescu-niculescu-mizil-2019-trouble}
without further adaptation.
All three encoders achieve above-chance AUROC
on both datasets, reaching 0.643 on Wikipedia
and 0.561 on ChangeMyView
(Appendix Table~\ref{tab:cga_transfer}).
Despite the domain and task shifts, these results
suggest that \toolname{} captures partially
transferable signals of conversational deterioration.


\begin{table}[t]
\centering
\small
\setlength{\tabcolsep}{4pt}
\resizebox{\columnwidth}{!}{%
\begin{tabular}{lcccc}
\toprule
& \multicolumn{2}{c}{\textbf{ConVAWG Test}} & \multicolumn{2}{c}{\textbf{MentalManip Test}} \\
\cmidrule(lr){2-3}\cmidrule(lr){4-5}
\textbf{Training data} & \textbf{F1$_\text{mac}$} & \textbf{AUC} & \textbf{F1$_\text{mac}$} & \textbf{AUC} \\
\midrule
\multicolumn{5}{l}{\textit{Full real training data}} \\
MentalManip (MM)                     & 0.755 & 0.904 & 0.674 & \textbf{0.807} \\
\toolname{} $\rightarrow$ MM         & \textbf{0.875} & \textbf{0.984} & \textbf{0.688} & 0.806 \\
\midrule
\multicolumn{5}{l}{\textit{Low-resource: 500 real examples}} \\
MM$_{500}$                           & 0.793 & 0.822 & 0.674 & \textbf{0.792} \\
\toolname{} $\rightarrow$ MM$_{500}$ & \textbf{0.956} & \textbf{0.995} & \textbf{0.689} & 0.791 \\
\bottomrule
\end{tabular}%
}
\caption{Cross-corpus evaluation of manipulative conversation detection with and without \toolname{} augmentation. MM denotes MentalManip, and $\rightarrow$ indicates sequential training. }
\label{tab:transfer_matrix}
\end{table}

\paragraph{Cross-dataset evaluation and augmentation.}
We evaluate the alignment between \toolname{} and
MentalManip \cite{wang-etal-2024-mentalmanip}, a
human-annotated corpus of movie dialogues.
A detector trained exclusively on MentalManip achieves
0.755 macro-F1 on \toolname{}, with positive prediction
rates increasing monotonically across escalation levels.
Conversely, synthetic pre-training improves performance
on \toolname{} while maintaining comparable performance
on MentalManip. In the low-resource setting, macro-F1
increases from 0.793 to 0.956 on \toolname{} and from
0.674 to 0.689 on MentalManip
(Table~\ref{tab:transfer_matrix}).
These results suggest that \toolname{} provides
complementary supervision for conversational
abuse detection.


\section{Conclusion}
\label{sec:conclusion}

We introduced \toolname, a retrieval-grounded framework for generating controlled synthetic VAWG dialogues. By combining CPS-aligned scenario construction, DHR-grounded retrieval, hierarchical event planning, persona-conditioned role-play, and toxicity injection, \toolname{} models abuse as a relational and temporally escalating phenomenon rather than isolated toxic utterances. The resulting dataset contains over 6,000 dialogue events across 200 scenarios with rich metadata, supporting evaluation of escalation, toxicity, relationship dynamics, and domain fidelity. Human evaluation, LLM-as-Judge assessment, ablations, and downstream tasks show that \toolname{} improves controllability and domain realism while maintaining strong dialogue quality.

\section*{Limitations}
\label{sec:limitations}
The scope of \toolname{} is constrained by the availability of reliable crime-related resources, including official definitions, population statistics, and publicly accessible case materials. Most grounding sources used in this work are UK-specific, including CPS definitions, ONS statistics, and Domestic Homicide Review reports. Consequently, the generated scenarios reflect UK legal categories, reporting conventions, demographic patterns, support-service structures, and documented abuse trajectories. The dataset may therefore transfer imperfectly to jurisdictions with different legal definitions, institutional practices, or cultural understandings of VAWG.

The term ``violence'' follows the broad CPS definition adopted in this work, encompassing coercive, controlling, and economic abuse in addition to physical and sexual violence. However, \toolname{} captures such behaviours only when they produce a conversational trace. Verbal abuse is directly represented in message text, whereas offline or behavioural abuse, such as financial control, surveillance, isolation, and stalking, is primarily encoded in the hierarchical event graphs and associated metadata. These behaviours may enter the dialogues through plans, updates, disclosures, or retrospective references. Institutional processes and non-communicative actions that do not plausibly surface in conversation remain outside the intended scope of the dataset.

\section*{Ethical Considerations}
\label{sec:ethical-considerations}

\paragraph{Intended Use and Scope.}
\toolname{} is intended to support research on
conversation-level abuse in VAWG contexts, including
abusive-language detection, escalation modelling,
relationship inference, and the evaluation of
safety-oriented NLP systems. By providing a controlled
synthetic resource, this work aims to facilitate
research in a domain where access to real conversational
data is restricted by ethical, privacy, and legal
considerations. Potential beneficiaries include
researchers, online-safety organisations, and
public-sector research bodies. However, the generated
dialogues and their associated annotations should not
be treated as evidence of actual abusive behaviour or
as a substitute for professional judgement. The dataset
is not intended for operational risk assessment or
decision-making concerning real individuals,
relationships, or legal cases.

\paragraph{Data Provenance and Privacy.}
No private conversations, victim messages, or
identifiable personal communications are included in
the released dataset. All dialogues are fictional
and generated from structured synthetic personas,
official crime definitions, aggregate demographic
statistics, and high-level patterns derived from
publicly available Domestic Homicide Reviews (DHRs).
These sources inform the generation of plausible
abuse dynamics rather than the reproduction or
reconstruction of individual cases. We recognise
that the public availability of such materials
does not eliminate ethical concerns surrounding
their use, particularly with respect to the dignity
and privacy of victims, survivors, and their families.

\paragraph{Human Annotation and Participant Welfare.}
During the human annotation study, all participants received a Participant Information
Sheet and provided informed consent prior to
participation. Given the potentially distressing
nature of the annotation material, several
safeguarding measures were implemented, including
an online debriefing procedure, the option to
discontinue participation or request contact with
the research team at any time, and access to contact
information for trained support professionals
affiliated with a partner charity.
Details of participant recruitment, compensation,
and annotation procedures are provided in
Appendix~\ref{human_annotation}.

\paragraph{Representational and Annotation Bias.}
Although \toolname{} is grounded in documented abuse
patterns and demographic statistics, synthetic
generation may reproduce or amplify stereotypes
concerning victims, perpetrators, and relationship
dynamics inherited from the underlying language
models and grounding sources. Moreover, perceptions
of abusive behaviour and conversational realism may
vary according to individual, cultural, and social
backgrounds. While the recruitment strategy was
informed by the gendered nature of VAWG and
consultation with our project partners, the human
evaluations reflect the perspectives of a limited
annotator population and should not be interpreted
as universally representative of VAWG experiences.

\paragraph{Dual-Use Risks and Responsible Release.}
The controlled generation of abusive and potentially
distressing language introduces inherent dual-use
risks. Although the proposed framework is designed
to advance abuse detection, analysis, and safety
evaluation, its generated content and methodology
could be misused to facilitate harmful communication
or develop systems that generate abusive language.
To mitigate these risks, access to the archival
dataset is governed by a data use agreement that
restricts use to non-commercial research and
prohibits harassment, profiling of real individuals,
training systems primarily intended to generate
abusive content, and application to real-world cases.
Access may be revoked following violations of these
conditions.

The calibrated toxicity steering vectors, associated
configurations, and toxicity-injection prompts are
excluded from the public release and may be made
available only upon justified request.
Released files include content warnings,
documentation of intended and prohibited uses,
and machine-readable metadata identifying the
data as synthetic. While these measures cannot
eliminate the possibility of misuse, they aim to
mitigate the risks associated with releasing
sensitive synthetic data while preserving its
scientific utility for responsible NLP research.

\section*{Acknowledgements}

XT is supported by the EPSRC [grant number EP/Y009800/1], through funding from Responsible AI UK (KP0016) as a Keystone project. This work made use of the Scientific Computing Research Technology Platform (SCRTP) at the University of Warwick for access to the Avon HPC cluster, and the Sulis Tier 2 HPC platform at HPC Midlands+, funded by the Engineering and Physical Sciences Research Council (EPSRC) under grant EP/T022108/1. This project is partially supported by the Police STAR Fund through the projects ``A Case Study of the Application of Natural Language Processing for Information Extraction and Analysis in Policy Review'' (23/24), conducted with the Forensic Capability Network (FCN), and ``STARIST: Stalking Threat AI Recognition (and) Identification Support Tool'' (25/26). It is also supported in part by the ESRC Digital Good Research Fund through the project "Making Harm Visible: Survivor-Centred Analysis of Domestic Homicide Reviews Using AI". The views expressed are those of the authors and do not necessarily reflect those of the FCN, the NPCC, or the OPCSA.

\bibliography{custom}

\appendix

\input{appendix}

\input{appendix_prompts}

\end{document}

%% file: appendix.tex
\section{Stage 1: Retrieval Grounding and Scenario Specification}
\label{app:stage1_details}

\paragraph{RAG on Domestic Homicide Reviews.}
\label{DHR_RAG}
Each raw DHR narrative is transformed by an LLM-based information-extraction pipeline into a structured case record (actors, relationships, an event timeline, a crime typology, and a factual case summary), which is embedded and indexed to support retrieval-augmented persona and scenario generation.

\paragraph{Structured persona and scenario output.}
\label{app:stage2_schema}
The persona and scenario generation step produces a structured scenario specification (victim profile, perpetrator profile, crime types, abuse duration, and optional third-party actors) that serves as the high-level blueprint for downstream event planning and dialogue generation. Two additional post-processing checks further enforce persona-scenario consistency and character-name diversity.

\textit{The full record and output schemas, the extractor configuration, the indexing and retrieval details, and the consistency-refinement procedure are provided in the project repository.}

\section{Stage 3: Language Style Control}
\label{sec:stage5_style}

To produce linguistically diverse and contextually appropriate conversations, we implement a retrieval-augmented style conditioning mechanism. Instead of encoding all stylistic rules directly in the prompt, the system retrieves relevant \emph{style notes} and behavioural cues from a structured knowledge base based on event attributes and persona profiles, and injects them into the dialogue generation prompt. This retrieval-conditioned prompting enables flexible control of linguistic style while maintaining consistency across scenarios.

\paragraph{Persona-conditioned informal language style.}
For each participant, the system retrieves messaging style notes derived from persona attributes, particularly age and demographic background. These notes describe the expected communication register (e.g., slang usage, abbreviation frequency, or formality level) and are inserted into the prompt before dialogue generation. Because style notes are retrieved independently for each role, speakers within the same conversation may exhibit distinct linguistic registers that reflect their persona characteristics.

\paragraph{Escalation-conditioned interaction style.}
The escalation level associated with the underlying event is also used as a retrieval key for selecting behavioural style notes. As escalation increases, the retrieved instructions progressively shift perpetrator language from rapport-building or subtle control toward manipulation, intimidation, or explicit threats. Correspondingly, victim responses transition from neutral or uncertain interaction to increasingly anxious, defensive, or distressed communication patterns. This escalation-aware retrieval ensures that conversational tone evolves consistently with the narrative dynamics.

\paragraph{Crime-type behavioural cues.}
Finally, the system retrieves crime-type-specific behavioural indicators derived from the event graph expansion stage (e.g., surveillance references in stalking scenarios or coercive monitoring in controlling relationships). These cues are injected as structured prompt guidance so that the language model incorporates realistic behavioural signals associated with the specific abuse modality rather than producing generic conflict dialogue.

\paragraph{Third-party conversations.}
When an event involves actors other than the perpetrator and victim (e.g., victim–friend interactions), the retrieval mechanism selects a dedicated set of style notes appropriate for third-party interactions. In these cases the prompt emphasizes the relational role of the third party—such as providing support, acting as a witness, or reinforcing isolation—rather than abusive escalation dynamics.

\subsection{Escalation-aware Control}
\label{app:escalation}
Each sub-event carries a discrete escalation level \(e \in \{0,1,2,3,4\}\): normal interaction (0), emerging control (1), manipulation and monitoring (2), coercive intimidation (3), and severe threats or violent escalation (4). The level is assigned by prompting the model with the event description, the escalation rubric, and the recent escalation history, and by validating the returned level against this range. It then acts as a global control signal across the pipeline: higher levels lead to finer-grained event decomposition, denser temporal spacing, longer and more confrontational exchanges, and the escalation-conditioned style notes described above. To reflect the tension-building, incident, reconciliation, and calm phases of intimate partner violence cycles, events that follow a high-escalation event (levels 3-4) may be assigned lower levels representing guilt, remorse, or ``honeymoon'' periods, which produces non-monotonic escalation trajectories.

\section{Toxicity Steering Details}
\label{app:toxicity_injection}

We use Contrastive Activation Addition (CAA; \citealt{rimsky-etal-2024-steering}) to steer
decoder-only language models toward toxic output at a controlled level of intensity.
At inference time a single forward hook is registered on a target transformer layer
$\ell^{*}$, perturbing the hidden state of every token position as
\begin{equation}
  h' = h + \alpha \cdot \hat{v}_{\ell^{*}} \label{eq:caa}
\end{equation}
where $h \in \mathbb{R}^{d}$ is the original hidden state,
$\hat{v}_{\ell^{*}}$ is the L2-normalised steering vector for layer $\ell^{*}$
with $\|\hat{v}_{\ell^{*}}\|_{2} = 1$,
and $\alpha \in \mathbb{R}_{\geq 0}$ is the scalar steering coefficient.
Because the vector direction is fixed to unit norm after normalisation,
$\alpha$ is the sole magnitude control and must be calibrated per model.

\paragraph{Vector Construction and Calibration.}
Steering vectors are derived from toxic-clean contrastive pairs sampled from the
Google Civil Comments dataset~\citep{borkan2019nuanced}: for a given layer, the raw
CAA vector is the mean hidden-state difference between the toxic and clean text of
each pair, subsequently L2-normalised. For each model we run a layer sweep to select
the steering layer $\ell^{*}$ and then calibrate three operating points (L1-L3)
spanning low, moderate, and high toxicity while preserving generation coherence,
evaluated on \texttt{RealToxicityPrompts}~\cite{gehman-etal-2020-realtoxicityprompts}
with Detoxify~\citep{hanu2020detoxify} toxicity together with perplexity and
diversity checks.

\textit{The exact pair-construction constants, the vector-extraction template and
hidden-state indexing, the layer-selection and calibration procedures, and the
per-model configuration (backbone models, selected steering layer, and the
level-specific coefficients $(\alpha_1,\alpha_2,\alpha_3)$ with their sweep and
calibration tables) form the calibrated steering configuration, which is not
included in the public release and is available upon justified request (see the
Ethical Considerations section).}
\label{app:caa:models}

\section{Evaluation Metrics}
\label{app:evaluation}
\paragraph{Dialogue-Level Metrics.}
These metrics assess the quality of the dialogue at the interaction level.

\begin{itemize} 

\item \textbf{Coherence.}
Measures logical consistency and conversational flow across turns \cite{xu-etal-2024-reasoning}. 
Higher scores indicate that utterances appropriately respond to preceding turns, maintain topical continuity, and avoid contradictions or abrupt shifts.

\item \textbf{Humanlikeness.}
Evaluates the naturalness and authenticity of the dialogue \cite{bao-etal-2023-synthetic}. 
Annotators assess linguistic fluency, natural phrasing, emotional plausibility, and whether the interaction resembles realistic human communication in an online messaging context.

\item \textbf{Persona Consistency.}
Assesses whether speakers consistently adhere to their predefined persona profiles (e.g., age, role, temperament, and linguistic style) throughout the conversation \cite{yang-etal-2025-crafting, lu-etal-2024-large}. 
High scores indicate stable behavioral patterns and stylistic consistency across turns.

\item \textbf{Toxicity Realism.}
Evaluates the contextual plausibility of toxic or abusive language when present \cite{luong-etal-2024-realistic}. 
Rather than measuring toxicity severity, this metric assesses whether toxic expressions emerge at appropriate escalation points, progress in intensity realistically, and align with the relationship dynamics depicted in the dialogue.

\end{itemize}

\paragraph{Global-Level Metrics.}

These metrics evaluate the dialogue relative to domain knowledge and scenario-level inputs.

\begin{itemize}

\item \textbf{Crime Fidelity.}
Assesses whether the dialogue realistically reflects patterns characteristic of VAWG-related crimes, including coercive control dynamics, power imbalances, escalation trajectories, and victim–perpetrator interaction structures.

\item \textbf{Scenario Realism.}
Measures the degree to which the generated dialogue aligns with the provided scenario summary. 
High scores indicate that key scenario elements—such as relationship context, major events, escalation patterns, and outcomes—are faithfully reflected in the dialogue while maintaining plausible narrative progression.

\end{itemize}

\section{Human Annotation Study Details}
\label{human_annotation}

\subsection{Annotator Recruitment and Compensation}
\label{app:annotator_demographics}

The human evaluation was conducted by a pool of
six female annotators recruited through Prolific,
an online participant recruitment platform.
Following consultation with our project partners,
female annotators were selected given the gendered
nature of the harms examined in this study.
This recruitment strategy is consistent with
previous work on gender-based abusive language
annotation, including the Explainable Detection
of Online Sexism dataset \cite{edos_dataset}.
Demographic diversity within the annotator group
was encouraged through platform-based recruitment
filters.

The initial compensation was estimated at
\pounds8 per hour based on the anticipated task
completion time. Final payments were adjusted
according to the actual time spent on annotation,
resulting in an effective hourly compensation
rate of approximately \pounds9.70.
Ethical approval and participant safeguarding
measures are described in
Section~\ref{sec:ethical-considerations}.

\subsection{Annotation Assignment}
\label{app:annotation_assignment}

The six annotators first rated a shared, randomly sampled subset of 10 scenarios
(379 dialogues), which is used to measure inter-annotator agreement
(Section~\ref{result: annotator}). The remaining scenarios were then distributed
among the same six annotators: each scenario was assigned to one annotator, who
rated all of its dialogues and provided the scenario-level ratings. Human ratings
therefore cover the full dataset, with repeated ratings on the shared subset.

\subsection{Annotation Platform}
\label{app:platform}

We developed a custom web-based annotation platform
using Streamlit.\footnote{\url{https://streamlit.io}}
For each scenario, the platform presents the scenario summary, the persona
profiles, and the associated dialogues. Annotators first read the Participant
Information Sheet and confirm informed consent, then review the annotation
guidelines, and then rate their assigned scenarios in any order. Progress is saved
automatically every minute so that the task can be paused and resumed, and the
platform verifies that all items are complete before accepting a submission.

\subsection{Annotation Task}
\label{app:task}

Each scenario comprises a temporally ordered sequence of dialogues corresponding
to individual sub-events of a VAWG case (e.g., E1.1\textsubscript{pre},
E1.2\textsubscript{during}). Before rating, annotators are shown the scenario
summary and the persona information (victim, perpetrator, their relationship, and
any third-party participants). They rate every dialogue on the four dialogue-level
dimensions and, after reading all dialogues of a scenario, give the two
scenario-level ratings, all on a five-point Likert scale from 1 (very poor) to
5 (excellent). The dimensions are defined in Appendix~\ref{app:evaluation}.

\paragraph{Utterance-Level Flagging.}
In addition to the quality ratings, annotators
may flag individual utterances exhibiting clear
generation errors through a checkbox displayed
alongside each message. Annotators were instructed
to flag an utterance only if it:

\begin{itemize}[leftmargin=*]
  \item is incomprehensible or contains garbled
        or meaningless characters;
  \item is entirely out of context given the
        dialogue, scenario, or speaker personas; or
  \item introduces an abrupt and unmotivated
        topic shift that disrupts conversational
        coherence.
\end{itemize}

This mechanism complements the quality ratings
by enabling the identification of localised
generation errors that may not be fully reflected
in dialogue-level scores.

\subsection{Quality Control and Attention Checks}
\label{app:qc}

To assess annotator attentiveness, explicit
attention checks were embedded throughout
the annotation tasks. Each check presents an
on-screen instruction requiring the annotator
to identify a specified utterance:

\begin{quote}
\small
\textit{``\textbf{ATTENTION CHECK} --
Please flag the following utterance by clicking
the checkbox next to it:
`\textlangle{}quoted utterance\textrangle{}'.''}
\end{quote}

Annotators were required to locate and flag
the specified utterance. An attention check
was considered failed if the target utterance
was not flagged.

As part of the quality-control protocol,
annotators failing more than two attention
checks were subject to exclusion from the
subsequent analyses. Annotation completeness
was additionally verified by the platform
prior to submission.

\section{Statistical Significance and Judge Robustness}
\label{app:significance}

To verify that the differences in Table~\ref{tab:multi_judge_comparison} reflect systematic quality differences rather than sampling noise, we treat the persona as the unit of analysis: for each of the 50 matched personas, every system receives one score per metric (per-dialogue metrics are first averaged within the persona; global metrics are single ratings), averaged over the four calibrated judges, which is exactly the aggregation of Table~\ref{tab:multi_judge_comparison}. Ratings that a judge failed to return (refusals or unparseable outputs) are dropped, and each persona is averaged over the judges that rated it; the cross-judge means in Table~\ref{tab:multi_judge_comparison} can therefore differ slightly from the mean of the per-judge rows in Table~\ref{tab:appendix_additional_llm_judges}. All tests below are run on these matched persona-level scores, using non-parametric procedures appropriate for ordinal judge ratings.

\subsection{Significance Tests}

\paragraph{Omnibus test.}
A Friedman test (the non-parametric repeated-measures ANOVA) across the nine systems rejects the null hypothesis of equal quality on all three aggregates: Avg-D $\chi^2(8){=}317.2$, Avg-G $\chi^2(8){=}298.8$, and Avg $\chi^2(8){=}325.8$ (all $p{<}10^{-4}$), with Kendall's $W$ of 0.79, 0.75, and 0.81 respectively, indicating strong concordance across personas in how the systems rank.

\paragraph{Post-hoc pairwise tests.}
Table~\ref{tab:significance_tests} reports, for each baseline, the paired mean difference to \toolname{} with a 95\% bootstrap percentile confidence interval (10{,}000 resamples over personas), the two-sided paired Wilcoxon signed-rank $p$-value with Holm-Bonferroni correction across the eight baselines, and the matched-pairs rank-biserial correlation $r$ as effect size.
\toolname{} is significantly ahead of every baseline on both Avg and Avg-D (all Holm-corrected $p{<}10^{-4}$, $r$ between ${+}0.67$ and ${+}1.00$).
On Avg-G, consistent with the analysis in Section~\ref{sec:main_quality_comparison}, the two strongest single-pass baselines hold a significant advantage (GPT-5.2 $\Delta{=}{-}0.16$, Gemini $\Delta{=}{-}0.08$) driven by one-shot global ratings of compact dialogue sets, while every other baseline remains significantly below \toolname{}.

\paragraph{Per-metric tests.}
Applying the same Holm-corrected Wilcoxon procedure to each metric separately, \toolname{} is significantly ahead of all eight baselines on Humanlikeness and Toxicity Realism (all Holm-corrected $p{<}10^{-3}$); ahead of seven baselines on Coherence (Gemini is ahead by 0.03, $p{<}0.01$); ahead of six on Persona Consistency (statistically indistinguishable from DiaSynth and SPASM); ahead of six on Crime Fidelity (indistinguishable from GPT-5.2 and Gemini at the 5-point ceiling); and ahead of six on Scenario Realism, where only GPT-5.2 and Gemini score higher (the one-shot global-rating effect discussed in Section~\ref{sec:main_quality_comparison}).

\begin{table*}[!t]
\centering
\small
\setlength{\tabcolsep}{2pt}
\resizebox{\linewidth}{!}{%
\begin{tabular}{l|ccc|ccc|ccc}
\toprule
 & \multicolumn{3}{c|}{\textbf{Avg-D (dialogue-level)}} & \multicolumn{3}{c|}{\textbf{Avg-G (global)}} & \multicolumn{3}{c}{\textbf{Avg (all six metrics)}} \\
\textbf{Baseline} & $\Delta$ [95\% CI] & $p_{\mathrm{Holm}}$ & $r$ & $\Delta$ [95\% CI] & $p_{\mathrm{Holm}}$ & $r$ & $\Delta$ [95\% CI] & $p_{\mathrm{Holm}}$ & $r$ \\
\midrule
LLaMA-3.1-70B-Instruct & $+0.51$ {\scriptsize$[+0.46, +0.55]$} & $<\!10^{-4}$ & $+1.00$ & $+2.02$ {\scriptsize$[+1.84, +2.18]$} & $<\!10^{-4}$ & $+1.00$ & $+1.01$ {\scriptsize$[+0.95, +1.07]$} & $<\!10^{-4}$ & $+1.00$ \\
Qwen3.5-35B-A3B        & $+0.35$ {\scriptsize$[+0.31, +0.38]$} & $<\!10^{-4}$ & $+1.00$ & $+0.51$ {\scriptsize$[+0.39, +0.64]$} & $<\!10^{-4}$ & $+0.97$ & $+0.40$ {\scriptsize$[+0.35, +0.46]$} & $<\!10^{-4}$ & $+1.00$ \\
DeepSeek-V4-Flash      & $+0.31$ {\scriptsize$[+0.27, +0.35]$} & $<\!10^{-4}$ & $+1.00$ & $+0.12$ {\scriptsize$[+0.04, +0.20]$} & $0.008$ & $+0.48$ & $+0.25$ {\scriptsize$[+0.20, +0.29]$} & $<\!10^{-4}$ & $+0.94$ \\
Gemini-3.1-Flash-Lite  & $+0.18$ {\scriptsize$[+0.15, +0.20]$} & $<\!10^{-4}$ & $+1.00$ & $-0.08$ {\scriptsize$[-0.13, -0.02]$} & $0.012$ & $-0.43$ & $+0.09$ {\scriptsize$[+0.06, +0.12]$} & $<\!10^{-4}$ & $+0.75$ \\
GPT-5.2                & $+0.18$ {\scriptsize$[+0.15, +0.21]$} & $<\!10^{-4}$ & $+0.99$ & $-0.16$ {\scriptsize$[-0.20, -0.12]$} & $<\!10^{-4}$ & $-0.90$ & $+0.07$ {\scriptsize$[+0.04, +0.10]$} & $<\!10^{-4}$ & $+0.67$ \\
\midrule
DiaSynth (Qwen3.5)     & $+0.11$ {\scriptsize$[+0.09, +0.13]$} & $<\!10^{-4}$ & $+0.96$ & $+0.15$ {\scriptsize$[+0.10, +0.21]$} & $<\!10^{-4}$ & $+0.76$ & $+0.12$ {\scriptsize$[+0.10, +0.16]$} & $<\!10^{-4}$ & $+0.95$ \\
SynDG (Qwen3.5)        & $+0.74$ {\scriptsize$[+0.66, +0.82]$} & $<\!10^{-4}$ & $+1.00$ & $+0.51$ {\scriptsize$[+0.41, +0.63]$} & $<\!10^{-4}$ & $+0.99$ & $+0.66$ {\scriptsize$[+0.59, +0.74]$} & $<\!10^{-4}$ & $+1.00$ \\
SPASM (Qwen3.5)        & $+0.44$ {\scriptsize$[+0.39, +0.48]$} & $<\!10^{-4}$ & $+1.00$ & $+0.62$ {\scriptsize$[+0.49, +0.75]$} & $<\!10^{-4}$ & $+1.00$ & $+0.50$ {\scriptsize$[+0.43, +0.57]$} & $<\!10^{-4}$ & $+1.00$ \\
\bottomrule
\end{tabular}%
}%
\caption{Statistical comparison of \toolname{} against each baseline under the four-judge average on the 50 matched personas (persona = unit of analysis, matching the aggregation of Table~\ref{tab:multi_judge_comparison}).
$\Delta$: mean paired difference (\toolname{} $-$ baseline) with 95\% bootstrap percentile CI over personas (10{,}000 resamples); positive values favour \toolname{}.
$p_{\mathrm{Holm}}$: two-sided paired Wilcoxon signed-rank $p$-value, Holm-corrected across the eight baselines within each column group.
$r$: matched-pairs rank-biserial correlation (effect size; $|r|{>}0.5$ conventionally large).
\toolname{} is significantly ahead of every baseline on Avg-D and Avg; the two negative Avg-G entries reflect the one-shot global-rating leniency toward compact single-pass dialogue sets analysed in Section~\ref{sec:main_quality_comparison}.}
\label{tab:significance_tests}
\end{table*}

\subsection{Judge Robustness}

\paragraph{Per-judge robustness.}
Repeating the Avg-D comparison within each judge separately (Holm-corrected within judge), \toolname{}'s dialogue-level advantage is significant ($p{<}0.05$) in 29 of the 32 judge-baseline pairs.
The three exceptions (GPT-5.2 and DiaSynth under the DeepSeek judge, and DiaSynth under the Qwen judge) are statistically indistinguishable from \toolname{}, and no baseline is rated significantly above \toolname{} on Avg-D by any judge.
The main-table conclusion therefore does not depend on any single judge model.

\paragraph{Leave-one-judge-out robustness.}
Because the GPT-5.2 judge shares the generation backbone of \toolname{} (and of the strongest direct baseline), and LLM evaluators are known to favour their own generations \cite{panickssery2024llm}, we additionally recompute the full Table~\ref{tab:multi_judge_comparison} aggregation with each judge held out in turn.
Excluding the GPT-5.2 judge, the remaining three-judge average ranks \toolname{} first on both aggregates (Avg 4.85 vs.\ 4.75 for DiaSynth, the strongest baseline; Avg-D 4.84 vs.\ 4.78), and every Holm-corrected Wilcoxon comparison against the eight baselines remains significant on both Avg and Avg-D (all $p{<}10^{-4}$, $r{\geq}{+}0.78$); removing the backbone judge thus widens rather than narrows \toolname{}'s margin, consistent with the self-preference direction documented in the global-metric analysis below.
\toolname{} ranks first on Avg-D under all four three-judge pools.
The only pool in which any baseline overtakes \toolname{} on the six-metric Avg is the one that retains the GPT-5.2 judge (holding out Qwen), where GPT-5.2 direct edges ahead by 0.02 (Holm-corrected $p{=}0.027$) through its one-shot global ratings; that is, the sole configuration favouring a baseline is the one containing the judge exposed to self-preference.

\paragraph{Per-judge global-metric gap.}
The per-judge breakdown in Table~\ref{tab:appendix_additional_llm_judges} locates the Scenario Realism deficit discussed in Section~\ref{sec:main_quality_comparison} in two judges: the GPT-5.2 judge assigns \toolname{} a constant rating of 4 across all 50 scenarios while scoring its own direct generations 4.88, and the DeepSeek judge rates \toolname{} 4.63 vs.\ 4.97; under the Gemini judge both systems sit at the ceiling (4.98 vs.\ 5.00), and under the Qwen judge \toolname{} is slightly ahead (4.58 vs.\ 4.56).
Together with the human reversal on the shared subset (Crime Fidelity 4.75 vs.\ 3.80, Scenario Realism 4.50 vs.\ 4.30), this indicates that the one-shot global-rating advantage of compact single-pass baselines reflects rating leniency toward brief scenario paraphrases rather than genuinely higher realism; \toolname{} itself stays within 0.01 of the Crime Fidelity ceiling (4.99).

\begin{table*}[!t]
\centering
\small
\setlength{\tabcolsep}{4pt}
\begin{tabular}{l|cccc|c|cc|c}
\toprule
\textbf{Generation Model} & \textbf{Coh.} & \textbf{Human.} & \textbf{Persona.} & \textbf{Toxic Real.} & \textbf{Avg-D} & \textbf{Crime.} & \textbf{Scenario.} & \textbf{Avg} \\
\midrule
\multicolumn{9}{c}{\textit{LLM Judge: GPT-5.2}} \\
\midrule
LLaMA-3.1-70B-Instruct & 4.77 & 3.89 & 3.46 & 3.99 & 4.03 & 3.32 & 3.06 & 3.75 \\
Qwen3.5-35B-A3B        & 4.70 & 3.90 & 3.67 & 4.36 & 4.16 & 4.64 & 3.94 & 4.20 \\
DeepSeek-V4-Flash      & 4.85 & 3.98 & 3.96 & 4.41 & 4.30 & 4.94 & 4.30 & 4.41 \\
Gemini-3.1-Flash-Lite  & \textbf{4.93} & 4.03 & 3.92 & 4.54 & 4.36 & 4.98 & 4.60 & 4.50 \\
GPT-5.2                & 4.87 & \textbf{4.04} & 3.90 & 4.65 & 4.37 & \textbf{5.00} & \textbf{4.88} & \textbf{4.56} \\
DiaSynth (Qwen3.5)     & 4.37 & 3.89 & 4.09 & 4.64 & 4.25 & 4.82 & 4.00 & 4.30 \\
SynDG (Qwen3.5)        & 3.94 & 2.78 & 3.79 & 3.82 & 3.58 & 4.02 & 3.90 & 3.71 \\
SPASM (Qwen3.5)        & 4.01 & 2.95 & 4.15 & 3.85 & 3.74 & 3.98 & 3.48 & 3.74 \\
\textbf{\toolname{}}   & 4.84 & 4.02 & \textbf{4.30} & \textbf{4.81} & \textbf{4.49} & \textbf{5.00} & 4.00 & 4.50 \\
\midrule
\multicolumn{9}{c}{\textit{LLM Judge: DeepSeek-V4-Flash}} \\
\midrule
LLaMA-3.1-70B-Instruct & 4.70 & 4.52 & 4.47 & 4.68 & 4.59 & 3.41 & 2.91 & 4.18 \\
Qwen3.5-35B-A3B        & 4.78 & 4.59 & 4.63 & 4.86 & 4.71 & 4.67 & 4.33 & 4.66 \\
DeepSeek-V4-Flash      & 4.79 & 4.61 & 4.65 & 4.84 & 4.72 & 4.87 & 4.84 & 4.76 \\
Gemini-3.1-Flash-Lite  & 4.79 & 4.64 & 4.68 & 4.86 & 4.74 & 4.97 & 4.94 & 4.80 \\
GPT-5.2                & 4.83 & 4.75 & 4.79 & 4.92 & 4.82 & \textbf{5.00} & \textbf{4.97} & 4.86 \\
DiaSynth (Qwen3.5)     & \textbf{4.90} & 4.72 & \textbf{4.88} & \textbf{4.99} & \textbf{4.87} & 4.94 & 4.74 & \textbf{4.87} \\
SynDG (Qwen3.5)        & 4.74 & 4.28 & 4.85 & 4.97 & 4.71 & 4.85 & 4.72 & 4.73 \\
SPASM (Qwen3.5)        & 4.78 & 4.36 & 4.84 & 4.95 & 4.73 & 4.83 & 4.50 & 4.71 \\
\textbf{\toolname{}}   & 4.84 & \textbf{4.78} & 4.84 & 4.96 & 4.85 & 4.89 & 4.63 & 4.83 \\
\midrule
\multicolumn{9}{c}{\textit{LLM Judge: Gemini-3.1-Flash-Lite}} \\
\midrule
LLaMA-3.1-70B-Instruct & 4.92 & 4.24 & 4.13 & 4.81 & 4.52 & 2.34 & 1.66 & 3.68 \\
Qwen3.5-35B-A3B        & 4.92 & 4.27 & 4.51 & 4.92 & 4.66 & 4.20 & 3.74 & 4.43 \\
DeepSeek-V4-Flash      & 4.89 & 4.46 & 4.57 & 4.93 & 4.71 & 4.70 & 4.70 & 4.71 \\
Gemini-3.1-Flash-Lite  & \textbf{5.00} & 4.78 & 4.73 & 4.99 & 4.87 & 4.98 & 4.96 & 4.91 \\
GPT-5.2                & 4.90 & 4.88 & 4.76 & \textbf{4.99} & 4.88 & \textbf{5.00} & \textbf{5.00} & 4.92 \\
DiaSynth (Qwen3.5)     & 4.92 & 4.34 & 4.95 & 4.97 & 4.79 & 4.83 & 4.81 & 4.80 \\
SynDG (Qwen3.5)        & 3.76 & 2.88 & 4.28 & 4.28 & 3.80 & 3.92 & 3.86 & 3.83 \\
SPASM (Qwen3.5)        & 4.69 & 3.86 & 4.82 & 4.84 & 4.55 & 4.39 & 4.28 & 4.48 \\
\textbf{\toolname{}}   & 4.99 & \textbf{4.98} & \textbf{4.95} & 4.97 & \textbf{4.97} & \textbf{5.00} & 4.98 & \textbf{4.98} \\
\midrule
\multicolumn{9}{c}{\textit{LLM Judge: Qwen3.5-35B-A3B}} \\
\midrule
LLaMA-3.1-70B-Instruct & 4.80 & 3.15 & 3.15 & 4.25 & 3.84 & 3.06 & 2.28 & 3.45 \\
Qwen3.5-35B-A3B        & 4.85 & 3.18 & 3.69 & 4.65 & 4.09 & 4.98 & 3.70 & 4.17 \\
DeepSeek-V4-Flash      & 4.75 & 3.24 & 3.66 & 4.55 & 4.05 & \textbf{5.00} & 3.90 & 4.18 \\
Gemini-3.1-Flash-Lite  & 4.95 & 3.54 & 4.11 & 4.75 & 4.33 & \textbf{5.00} & 4.34 & 4.45 \\
GPT-5.2                & 4.61 & 3.71 & 3.73 & 4.75 & 4.20 & \textbf{5.00} & 4.56 & 4.39 \\
DiaSynth (Qwen3.5)     & \textbf{4.95} & 3.89 & \textbf{4.89} & \textbf{4.99} & 4.68 & \textbf{5.00} & 4.00 & 4.62 \\
SynDG (Qwen3.5)        & 4.41 & 2.49 & 4.21 & 4.73 & 3.96 & 4.88 & 4.12 & 4.14 \\
SPASM (Qwen3.5)        & 4.86 & 2.92 & 4.57 & 4.96 & 4.33 & 4.78 & 3.52 & 4.27 \\
\textbf{\toolname{}}   & 4.87 & \textbf{4.27} & 4.64 & 4.96 & \textbf{4.69} & \textbf{5.00} & \textbf{4.58} & \textbf{4.72} \\
\bottomrule
\end{tabular}
\caption{
Per-judge LLM-as-Judge results underlying the four-judge average in Table~\ref{tab:multi_judge_comparison}.
Each block reports one judge model rating the same generation systems on the 50 matched personas; columns and aggregation are as in the LLM-as-Judge block of Table~\ref{tab:multi_judge_comparison}.
\textbf{Bold} marks the best score per column within each judge block.
Each row averages over the personas for which that judge returned a rating; because missing ratings are dropped, the mean of the four blocks need not equal Table~\ref{tab:multi_judge_comparison} exactly.
All quality scores are reported on a 1-5 scale.
}
\label{tab:appendix_additional_llm_judges}
\end{table*}

\subsection{Score Distributions and Ceiling Effects}

\begin{figure*}[t]
\centering
\includegraphics[width=\textwidth]{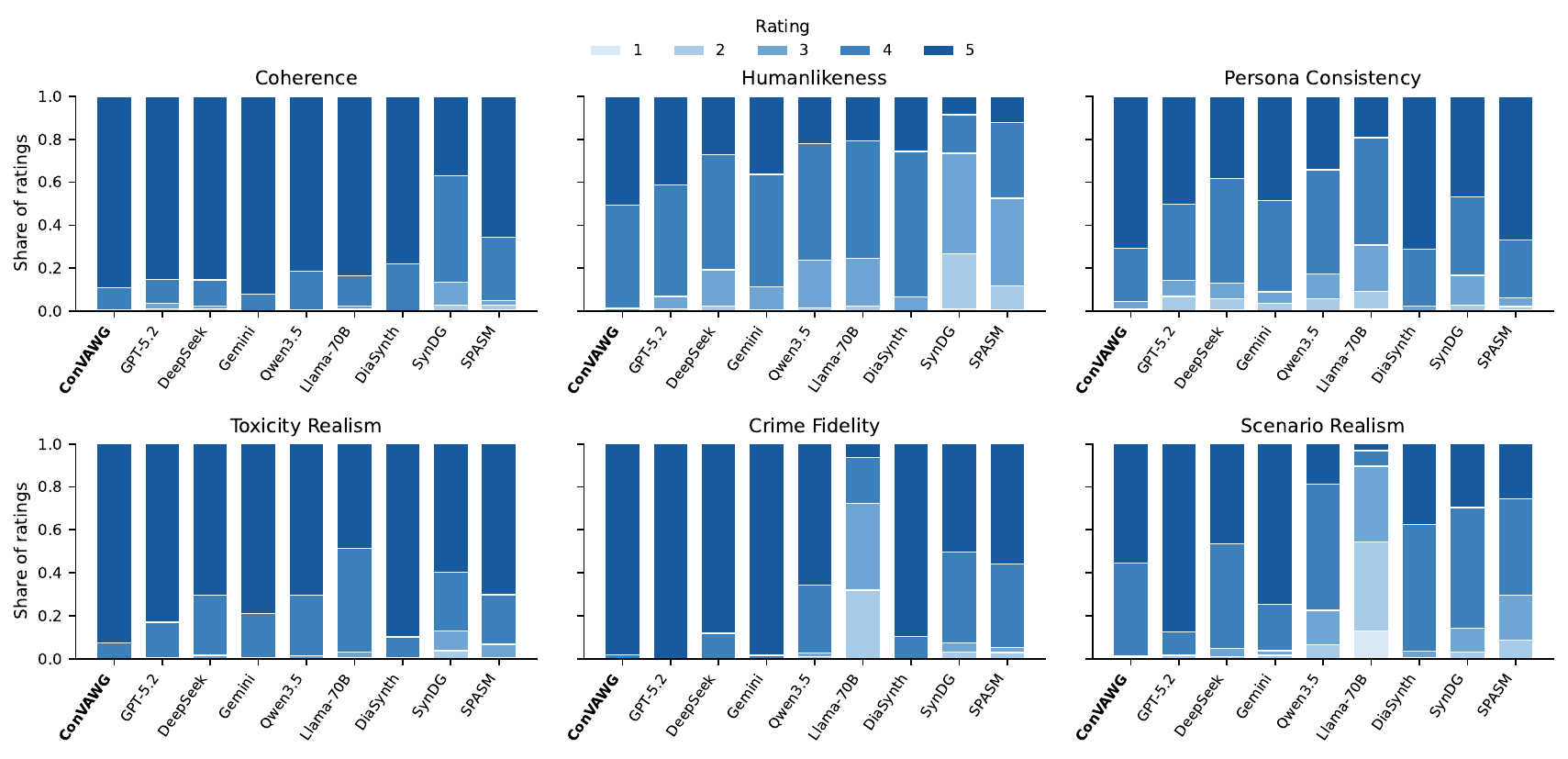}
\caption{Distribution of LLM-judge ratings (1-5) per evaluation metric, pooled over the four calibrated judges on the 50 matched personas, for all nine generation systems. Ratings concentrate at the top of the scale (57.5\% of all pooled ratings are 5), but the share of top ratings still separates systems on the discriminative metrics.}
\label{fig:score_distributions}
\end{figure*}

Judge ratings concentrate near the top of the 1-5 scale, a known property of LLM-as-Judge scoring \cite{zheng2023judging}: pooling all per-dialogue and global ratings from the four judges on the 50 matched personas (171{,}155 ratings), 57.5\% are exactly 5 and 89.3\% are 4 or above (for \toolname{}: 75.1\% and 98.3\%).
Following recommended reporting practice for LLM-as-Judge evaluations \cite{lee2026reporting}, Figure~\ref{fig:score_distributions} therefore reports the full per-metric rating distributions of all nine systems rather than means alone.
Despite the compression, the distributions still separate systems: on Humanlikeness, for example, \toolname{} receives a rating of 5 for 50.7\% of dialogues vs.\ 41.4\% for direct GPT-5.2 and 27.1\% for DeepSeek-V4. All conclusions in this appendix rest on paired persona-level comparisons and rank-based tests rather than absolute means.
Mean scores this close to the ceiling nevertheless compress the visible differences between strong systems.

\section{Additional Baseline Comparison Statistics}
\label{app:baseline_stats}

Table~\ref{tab:scale_comparison} reports the size of each system's output on the 50 matched personas, and Table~\ref{tab:toxic_rate} reports the mean Detoxify toxicity of the same outputs; both are discussed in Section~\ref{sec:main_quality_comparison}.

\begin{table}[t]
\centering
\small
\resizebox{\columnwidth}{!}{%
\begin{tabular}{lccc}
\toprule
\textbf{System} & \textbf{Avg Dialogues / Scenario} & \textbf{Avg Turns / Dialogue} & \textbf{Avg Words / Dialogue} \\
\midrule
LLaMA-3.1-70B-Instruct & 22.1 &  2.7 &   37.7 \\
Qwen3.5-35B-A3B        & 22.0 &  5.7 &   74.5 \\
DeepSeek-V4-Flash      & 24.8 &  4.2 &   51.6 \\
Gemini-3.1-Flash-Lite  & 20.0 &  2.8 &   38.4 \\
GPT-5.2                & 24.3 &  6.8 &   67.6 \\
\midrule
DiaSynth (Qwen3.5)     & 24.6 & 18.9 &  397.4 \\
SynDG (Qwen3.5)        & 24.6 & 16.0 & 1284.3 \\
SPASM (Qwen3.5)        & 24.6 & 15.2 & 1356.7 \\
\midrule
\textbf{\toolname{}}   & 34.2 & 14.5 &  200.1 \\
\bottomrule
\end{tabular}%
}
\caption{Scale comparison between \toolname{} and the eight baselines on the 50 matched personas: average number of dialogues per scenario, turns per dialogue, and words per dialogue (whitespace tokens of the rendered \textit{Speaker: text} transcript). Over the full 200-scenario release, \toolname{} averages 30.9 dialogues per scenario, 14.5 turns and 201.6 words per dialogue.}
\label{tab:scale_comparison}
\end{table}

\begin{table}[t]
\centering
\small
\setlength{\tabcolsep}{5pt}
\begin{tabular}{l c}
\toprule
\textbf{Generation Model} & \textbf{Toxic Rate} \\
\midrule
LLaMA-3.1-70B-Instruct & 0.07 \\
Qwen3.5-35B-A3B & 0.14 \\
DeepSeek-V4-Flash & 0.10 \\
Gemini-3.1-Flash-Lite & 0.10 \\
GPT-5.2 & 0.08 \\
\midrule
DiaSynth (Qwen3.5) & 0.15 \\
SynDG (Qwen3.5) & 0.07 \\
SPASM (Qwen3.5) & 0.15 \\
\midrule
\textbf{\toolname{}} & \textbf{0.17} \\
\bottomrule
\end{tabular}
\caption{
Mean Detoxify \cite{hanu2020detoxify} toxicity scores across generation systems on the 50 matched personas. Higher values indicate greater detected toxic content.
}
\label{tab:toxic_rate}
\end{table}

\section{Ablation Details}
\label{app:ablation}

Tables~\ref{tab:backbone_ablation_judge} and~\ref{tab:steering_compact} give the full results of the backbone and toxic-steering ablations discussed in the main text.

\begin{table*}[!t]
\centering
\small
\setlength{\tabcolsep}{4pt}
\begin{tabular}{l|cccccc|c}
\toprule
\textbf{Generation Model} & \textbf{Coh.} & \textbf{Human.} & \textbf{Persona.} & \textbf{Toxic Real.} & \textbf{Crime.} & \textbf{Scenario.} & \textbf{Avg} \\
\midrule
\multicolumn{8}{c}{\textit{Average over three judges}} \\
\midrule
  LLaMA-3.1-8B & 1.86 & 1.83 & 2.23 & 2.56 & 2.00 & 1.36 & 1.97 \\
  Qwen3.5-35B-A3B & 3.06 & 2.96 & 3.04 & 3.80 & 2.18 & 1.85 & 2.82 \\
  Gemini-3.1-Flash-Lite & \textbf{4.94} & \textbf{4.61} & 4.54 & \textbf{4.95} & 3.45 & 3.13 & 4.27 \\
  DeepSeek-V4-Flash & 4.86 & 4.58 & 4.42 & 4.70 & 4.70 & 4.07 & 4.56 \\
  \textbf{GPT-5.2}    & 4.89 & 4.59 & \textbf{4.70} & 4.91 & \textbf{4.96} & \textbf{4.54} & \textbf{4.77} \\
\midrule
\multicolumn{8}{c}{\textit{LLM Judge: GPT-5.2}} \\
\midrule
  LLaMA-3.1-8B & 1.66 & 1.59 & 2.05 & 2.45 & 2.14 & 1.29 & 1.86 \\
  Qwen3.5-35B-A3B & 1.95 & 1.94 & 1.93 & 3.15 & 2.38 & 1.88 & 2.20 \\
  Gemini-3.1-Flash-Lite & \textbf{4.88} & \textbf{4.12} & 4.00 & \textbf{5.00} & 1.40 & 1.30 & 3.45 \\
  DeepSeek-V4-Flash & 4.74 & 4.00 & 3.88 & 4.39 & 4.80 & 3.70 & 4.25 \\
  \textbf{GPT-5.2}    & 4.84 & 4.02 & \textbf{4.30} & 4.81 & \textbf{5.00} & \textbf{4.00} & \textbf{4.50} \\
\midrule
\multicolumn{8}{c}{\textit{LLM Judge: DeepSeek-V4-Flash}} \\
\midrule
  LLaMA-3.1-8B & 2.01 & 1.94 & 2.27 & 2.56 & 2.00 & 1.50 & 2.05 \\
  Qwen3.5-35B-A3B & 2.82 & 2.66 & 2.93 & 3.73 & 1.33 & 1.00 & 2.41 \\
  Gemini-3.1-Flash-Lite & \textbf{4.96} & \textbf{4.79} & 4.81 & 4.90 & 4.25 & 4.00 & 4.62 \\
  DeepSeek-V4-Flash & 4.88 & 4.78 & 4.76 & 4.86 & 4.75 & 4.50 & 4.76 \\
  \textbf{GPT-5.2}   & 4.84 & 4.78 & \textbf{4.84} & \textbf{4.96} & \textbf{4.89} & \textbf{4.63} & \textbf{4.83} \\
\midrule
\multicolumn{8}{c}{\textit{LLM Judge: Gemini-3.1-Flash-Lite}} \\
\midrule
  LLaMA-3.1-8B & 1.90 & 1.95 & 2.38 & 2.67 & 1.86 & 1.29 & 2.01 \\
  Qwen3.5-35B-A3B & 4.42 & 4.27 & 4.27 & 4.53 & 2.83 & 2.67 & 3.83 \\
  Gemini-3.1-Flash-Lite & 4.97 & 4.91 & 4.80 & 4.94 & 4.70 & 4.10 & 4.74 \\
  DeepSeek-V4-Flash & 4.95 & 4.97 & 4.63 & 4.84 & 4.56 & 4.00 & 4.66 \\
  \textbf{GPT-5.2}    & \textbf{4.99} & \textbf{4.98} & \textbf{4.95} & \textbf{4.97} & \textbf{5.00} & \textbf{4.98} & \textbf{4.98} \\
\bottomrule
\end{tabular}
\caption{Backbone model ablation study. Rows are different backbone models under the \toolname\ pipeline; the top block reports the mean over the three LLM judges, followed by the per-judge blocks. Columns are the same as in Table~\ref{tab:multi_judge_comparison}. The \textbf{GPT-5.2} row is the production \toolname{} configuration, reported from the per-judge results on the 50 matched personas (Table~\ref{tab:appendix_additional_llm_judges}).}
\label{tab:backbone_ablation_judge}
\end{table*}

\begin{table}[t]
\centering
\scriptsize
\setlength{\tabcolsep}{4pt}
\renewcommand{\arraystretch}{0.9}
\begin{tabular}{lcccc}
\toprule
\textbf{Condition} & \textbf{GPT} & \textbf{DeepSeek} & \textbf{Gemini} & \textbf{Toxic Rate} \\
\midrule
\textsc{Full}       & \textbf{4.62} & \textbf{4.83} & 4.96          & 0.144 \\
\textsc{w/o Steer.} & 4.54          & 4.75          & \textbf{4.97} & \textbf{0.038} \\
\bottomrule
\end{tabular}
\caption{
Ablation on disabling CAA toxic steering: overall LLM-judge average under each judge model (GPT-5.2, DeepSeek-V4, Gemini-3.1) and mean Detoxify toxicity (Toxic Rate).
}
\label{tab:steering_compact}
\end{table}

\section{Downstream Task Details}
\label{app:downstream}

This appendix gives the task definitions, splits, and protocols for the downstream experiments of Section~\ref{sec:downstream}, together with the results that are not shown in the main text.

\begin{table}[t]
\centering
\small
\setlength{\tabcolsep}{4pt}
\begin{tabular}{lcccc}
\toprule
\textbf{Model} & \textbf{W-F1} & \textbf{Acc.} & \textbf{Prec.} & \textbf{Rec.} \\
\midrule
BERT-base       & 0.629 & 0.636 & 0.339 & \textbf{0.346} \\
RoBERTa-base    & 0.530 & 0.541 & 0.279 & 0.288 \\
Longformer-4096 & 0.580 & 0.613 & 0.304 & 0.333 \\
GPT-5.2         & 0.523 & 0.525 & \textbf{0.346} & 0.271 \\
DeepSeek        & 0.360 & 0.524 & 0.131 & 0.250 \\
Gemini          & \textbf{0.665} & \textbf{0.651} & 0.341 & 0.318 \\
\bottomrule
\end{tabular}
\caption{Perpetrator-victim relationship classification results. W-F1 denotes weighted F1.}
\label{tab:relation_clf}
\end{table}

\paragraph{Toxic behaviour classification.}
A message-level eight-class task over the behaviour labels attached during toxicity injection: six perpetrator behaviours (\textit{surveillance}, \textit{demand}, \textit{accusation}, \textit{manipulation}, \textit{ultimatum}, \textit{threat}) plus \textit{non-toxic} and a merged \textit{other} class. Each example is a single message: S1 presents the message alone, and S2 adds the preceding dialogue as context. We fine-tune BERT-base, RoBERTa-base, and DeBERTa-v3-base \citep{tan-etal-2025-safespeech,khan-etal-2025-multilingual,alajmi-pergola-2025-leveraging}, and compare them with zero-shot GPT-5.2, DeepSeek, and Gemini (Table~\ref{tab:multiclass_classification} in the main text).

\paragraph{Relationship classification.}
A five-class perpetrator-victim relationship task (intimate partner, ex-partner, acquaintance, professional, and family/other), with the label taken from the persona metadata of each scenario. BERT-base, RoBERTa-base, and the zero-shot LLMs \citep{sun-etal-2024-leveraging,sun-etal-2026-boundary} classify one event-level dialogue at a time, whereas Longformer-4096 receives the dialogues of the whole scenario. Results are reported in Table~\ref{tab:relation_clf}; label imbalance and subtle contextual differences between relationship types make this task harder than behaviour classification.

\paragraph{Escalation forecasting.}
The benchmark uses a scenario-level 70/15/15 split (3{,}329/673/716 next-conversation examples over 140/30/30 scenarios); inputs are the raw text of all prior conversations in the scenario, with no escalation annotations. Every system predicts the escalation level of the next conversation, and the jump task is derived from these level predictions rather than trained separately: an example is a gold jump when the next level exceeds the current level (135 of the 716 test examples, 18.9\%), a system predicts a jump when its predicted level exceeds the current level, and F1$_\text{jump}$ is the F1 of the jump class. We include three trivial baselines. \emph{Persistence} copies the current level. \emph{Markov-1} predicts the most frequent next level given the current level, estimated from the transitions in the training split; it coincides with Persistence because the most likely next level equals the current level in every state. \emph{Majority} predicts, for every example, the most frequent next level in the training split (level~3, 55.4\% of training examples). Persistence therefore never predicts a jump, whereas Majority predicts a jump for every example whose current level is below~3, which gives it a non-zero F1$_\text{jump}$ despite a constant level prediction. We fine-tune DeBERTa-v3-base, Longformer-4096, and ModernBERT-base, and prompt GPT-5.2, DeepSeek-V4, and Qwen3.5 directly; for Qwen3.5 we also report generation-based forecasting \cite{zhang-etal-2025-forecasting}. Two observations complement Table~\ref{tab:escalation_forecast} in the main text: in an early-warning analysis, Longformer flags one third of first severe (level-4) events with zero false alerts, and under direct prompting GPT-5.2 and DeepSeek-V4 predict level~4 for roughly 80\% of examples.

\paragraph{Cross-corpus transfer protocol.}
\toolname-trained forecasters score derailment on the Conversations Gone Awry corpora (Wikipedia talk pages, $n = 837$; Reddit ChangeMyView, $n = 1{,}220$) as $P(\text{next level} \geq 3)$, under the standard forecasting protocol in which the attack comment is never part of the input; no CGA data is used for training. Both corpora are label-balanced, so chance AUROC is 0.5. Table~\ref{tab:cga_transfer} reports the AUROC of each forecaster.

\begin{table}[t]
\centering
\small
\setlength{\tabcolsep}{6pt}
\begin{tabular}{lcc}
\toprule
& \multicolumn{2}{c}{\textbf{AUROC\,$\uparrow$}} \\
\cmidrule(lr){2-3}
\textbf{Model} & \textbf{Wiki} & \textbf{CMV} \\
\midrule
DeBERTa-v3-base & 0.599 & \textbf{0.561} \\
Longformer-4096 & \textbf{0.643} & 0.556 \\
ModernBERT-base & 0.576 & 0.541 \\
\midrule
Chance & 0.500 & 0.500 \\
\bottomrule
\end{tabular}
\caption{Zero-shot transfer of \toolname{}-trained
escalation forecasters to conversational derailment
prediction on Wikipedia (Wiki) and ChangeMyView (CMV).
Bold indicates the best result per corpus.}
\label{tab:cga_transfer}
\end{table}

\paragraph{MentalManip alignment.}
MentalManip \cite{wang-etal-2024-mentalmanip} contains 2{,}915 dialogues with human consensus labels. The shared task is dialogue-level detection of manipulative or abusive conversation: \toolname{} positives are dialogues with escalation $\geq 2$, negatives include third-party support conversations as hard negatives, and speakers in both corpora are anonymised to remove formatting artefacts. Table~\ref{tab:transfer_matrix} reports DeBERTa-v3-base trained on real data alone or with \toolname{} pre-training, evaluated on both test sets; the per-level positive rates of the MentalManip-only detector on \toolname{} are 0.37/0.56/0.94/0.99/1.00 over escalation levels 0-4.

\paragraph{Implementation details.}
All \toolname{} splits are made at the scenario level (70/15/15, seed 42), so that no scenario contributes to more than one split. In the behaviour-classification, escalation-forecasting, and MentalManip experiments, encoders are fine-tuned for three epochs with a learning rate of $2\times10^{-5}$, and inputs are truncated to 512 tokens (4{,}096 for Longformer and ModernBERT); forecasting inputs are truncated from the left so that the most recent conversations are retained. All experiments were run on NVIDIA RTX 3090 GPUs.

%% file: appendix_prompts.tex
\raggedbottom
\onecolumn
\section{Pipeline Prompts}
\label{sec:appendix_prompts}

Table~\ref{tab:prompt_overview} summarises the prompts used by the generation
pipeline and the downstream tasks. The full prompt text for Stages~1--3 and for
the downstream tasks is provided in the project repository. The LLM-as-Judge
evaluation prompts (Appendix~\ref{app:stage6}) are reproduced in full, so that
the evaluation protocol can be independently inspected and reproduced. The
Stage~4 toxicity-injection prompts, style guidelines, and system prompt are
withheld from the public release to limit dual-use risk (see the Ethical
Considerations section) and may be provided only upon justified request.

\begin{table*}[t]
\centering
\small
\begin{tabularx}{\textwidth}{@{}l l X l@{}}
\toprule
\textbf{Stage} & \textbf{Prompt} & \textbf{Role} & \textbf{Availability} \\
\midrule
Stage 1 & Profile generation and scenario description & A system prompt establishes the expert persona; an instruction prompt requests a JSON-structured output with victim and perpetrator profiles, crime types, duration, and a narrative scenario description. & Repository \\
Stage 2 & Event graph extraction & Extracts a structured event graph of 5--8 composite events with directed relations from the completed scenario record. & Repository \\
Stage 2 & Event graph expansion & Decomposes each high-level event into 2--4 sub-events with locations, agents, and inter-event relations. & Repository \\
Stage 3 & Main dialogue generation & Converts each chat-script specification into a multi-turn online conversation. & Repository \\
Stage 3 & Conversation continuity block & Injected when a prior conversation exists, so that tone, topic, and relationship dynamics carry over according to the elapsed time gap. & Repository \\
Stage 3 & Third-party conversation & Template for conversations between the victim and a support contact (friend, family member, or professional). & Repository \\
Stage 3 & Perpetrator style notes & Rule-based notes keyed to escalation level and crime type, shifting language from rapport-building through subtle control to explicit threats, with a post-incident guilt phase. & Repository \\
Stage 3 & Victim style notes & Matching notes that shift responses from natural engagement through stress and self-doubt to fear and trauma responses, with a reconciliation phase. & Repository \\
Stage 4 & Toxicity injection & Rewrites a sampled subset of perpetrator messages labelled as escalation acts, with the steering strength set by the escalation level of the source event; victim messages are not modified. & On request \\
\midrule
Downstream & Behaviour classification & Classifies a single message into one of eight behaviour categories, from the message alone (S1) or with the preceding dialogue as context (S2). & Repository \\
Downstream & Relationship classification & Predicts the perpetrator--victim relationship type from five labels given a conversation. & Repository \\
\bottomrule
\end{tabularx}
\caption{Prompts used by the \toolname{} pipeline and the downstream tasks, with the role of each and where its full text is available.}
\label{tab:prompt_overview}
\end{table*}

\subsection{LLM-as-Judge Evaluation}
\label{app:stage6}

An LLM judge evaluates each generated scenario on six dimensions: four
per-dialogue metrics and two global metrics across all dialogues.

\paragraph{Judge System Prompt.}
\begin{Verbatim}[fontsize=\small, breaklines=true]
You are an expert evaluator for synthetic dialogue datasets about Violence
Against Women and Girls (VAWG).
Your task is to assess the quality of generated chat-based conversations on
specific rubric dimensions.
Respond ONLY with valid JSON -- no markdown fences, no extra text.
\end{Verbatim}

\paragraph{Per-Dialogue Rubric (4 metrics).}
\begin{Verbatim}[fontsize=\small, breaklines=true]
Rate the single dialogue below on FOUR metrics.

METRICS & GUIDANCE
==================

1. coherence
   Turn-level logical consistency and conversational flow.
   - Appropriate responses to previous utterances
   - Topic continuity
   - Absence of contradictions or abrupt shifts
   Judge internal conversational logic ONLY -- not scenario accuracy.

2. humanlikeness
   Naturalness and conversational authenticity.
   - Linguistic fluency, natural phrasing
   - Emotional appropriateness
   - Realistic conversational behaviour in online chatting context
   - Treat typical chat traits (short replies, mild typos, slang, imperfect
     grammar) as normal unless they significantly harm clarity
   Focus on language quality and interaction style -- not domain realism.

3. persona_consistency
   Stability of character traits and speaking style.
   - Tone, vocabulary, behavioural patterns
   - Emotional reactions consistent with assigned profile
   - Alignment with character description
   Consider both speakers across all turns.

4. toxicity_realism
   Realism of toxic / abusive language (if any).
   - Emerges at contextually appropriate points
   - Progresses in intensity in a believable way
   - Aligns with relationship dynamics and situation
   Judge REALISM of timing and expression only -- NOT morality or toxicity
   level.
   If no toxic language is present, rate "N/A" and leave explanation empty.

RATING SCALE (1-5)
==================
1 -- Very Poor:  Serious problems that substantially undermine realism,
                 consistency, or plausibility. Use only when issues are
                 frequent and clearly damaging.
2 -- Weak:       Noticeable issues that reduce quality or realism.
3 -- Moderate:   Acceptable overall with imperfections.
4 -- Strong:     Mostly realistic and consistent with only minor issues.
5 -- Excellent:  Highly realistic, coherent, and contextually appropriate.

IMPORTANT: Judge within context. Consider assigned personas and relationship
dynamics. Be fair and calibrated; do not over-penalize minor stylistic issues
common in chat. Reserve 1-2 for substantial problems.

OUTPUT FORMAT (JSON only, no markdown):
{
  "coherence":           {"rating": "<1|2|3|4|5>", "explanation": "<1-3 sentences>"},
  "humanlikeness":       {"rating": "<1|2|3|4|5>", "explanation": "<1-3 sentences>"},
  "persona_consistency": {"rating": "<1|2|3|4|5>", "explanation": "<1-3 sentences>"},
  "toxicity_realism":    {"rating": "<1|2|3|4|5|N/A>", "explanation": "<1-3 sentences>"}
}
\end{Verbatim}

\paragraph{Global Rubric (2 metrics across all dialogues).}
\begin{Verbatim}[fontsize=\small, breaklines=true]
After reading the FULL dialogue sequence and scenario summary, rate TWO
global metrics.

METRICS & GUIDANCE
==================

1. crime_fidelity
   Domain-level realism of VAWG dynamics across ALL dialogues.
   - Coercion and control behaviours
   - Escalation trajectories
   - Power imbalance
   - Victim-perpetrator interaction patterns
   Focus on domain authenticity -- not writing quality or scenario alignment.

2. scenario_realism
   Faithful and plausible alignment with the GIVEN SCENARIO across ALL
   dialogues.
   - Key elements reflected (relationship, ages, cultural background,
     vulnerabilities)
   - Logical consistency with described events
   - Realistic pacing and progression
   Judge alignment to this specific scenario -- not general VAWG realism.

RATING SCALE (1-5): same scale as per-dialogue rubric above.

Be fair and calibrated: avoid over-penalizing minor chat-style imperfections.
Reserve 1-2 for substantial failures.

OUTPUT FORMAT (JSON only, no markdown):
{
  "crime_fidelity":   {"rating": "<1|2|3|4|5>", "explanation": "<2-4 sentences>"},
  "scenario_realism": {"rating": "<1|2|3|4|5>", "explanation": "<2-4 sentences>"}
}
\end{Verbatim}

\clearpage
\subsection{Crime Type Taxonomy}
\label{app:crime_types}

Table~\ref{tab:crime_taxonomy} lists the 24 crime-type labels used throughout
the pipeline, together with a catch-all \texttt{other} label, their grouping,
and brief definitions derived from CPS VAWG strategy documentation. Of these, 18 occur in the released dataset.

\begin{table}[H]
\centering
\small
\setlength{\tabcolsep}{4pt}
\begin{tabular}{p{0.26\linewidth} p{0.18\linewidth} p{0.50\linewidth}}
\toprule
\textbf{Label} & \textbf{Category} & \textbf{Definition} \\
\midrule
\texttt{physical}        & Domestic abuse      & Physical assault, violence, bodily harm. Part of domestic abuse patterns. \\
\texttt{psychological}   & Domestic abuse      & Intimidation, threats, and humiliation that cause fear or distress. May include isolation. \\
\texttt{emotional}       & Domestic abuse      & Emotional abuse, degradation, and humiliation that damages self-worth. \\
\texttt{economic}        & Domestic abuse      & Controlling financial resources; preventing access to money or goods. \\
\texttt{financial}       & Domestic abuse      & Exploitation of property or goods for personal gain. \\
\texttt{coercive\_control} & Domestic abuse    & Pattern of controlling or coercive behaviour (Serious Crime Act 2015, \S76): isolation, deprivation, monitoring, regulating daily behaviour. \\
\midrule
\texttt{stalking}        & Stalking/harassment & Fixated, Obsessive, Unwanted, Repeated (FOUR) intrusive behaviour: following, unwanted contact, surveillance, threats. \\
\texttt{harassment}      & Stalking/harassment & Repeated behaviour ($\geq$2 occasions) that causes alarm or distress. \\
\texttt{cyberstalking}   & Stalking/harassment & FOUR stalking pattern conducted via digital means (social media, email, tracking apps). \\
\texttt{digital}         & Stalking/harassment & Online monitoring, surveillance, harassment, or use of technology to control. \\
\midrule
\texttt{sexual}          & Sexual offences     & General non-consensual sexual acts. \\
\texttt{rape}            & Sexual offences     & Non-consensual penile penetration (vagina, mouth, or anus); Sexual Offences Act 2003. \\
\texttt{sexual\_assault} & Sexual offences     & Intentional, non-consensual sexual touching. \\
\texttt{assault\_by\_penetration} & Sexual offences & Non-penile penetration of vagina or anus without consent. \\
\texttt{child\_sexual\_abuse} & Sexual offences & Sexual abuse of persons under 18, including grooming, exploitation, and contact/non-contact offences. \\
\midrule
\texttt{child\_abuse\_non\_sexual} & Child abuse & Non-sexual maltreatment of persons under 18: violence, cruelty, causing harm. \\
\texttt{neglect}         & Child abuse         & Failure to provide care or prevent harm to a child. \\
\midrule
\texttt{honor\_based}    & Honour-based abuse  & Violence or coercion to protect/defend perceived family or community honour. \\
\texttt{forced\_marriage}& Honour-based abuse  & Marriage without free and full consent; use of violence, threats, or pressure. \\
\texttt{fgm}             & Honour-based abuse  & Female Genital Mutilation: all WHO Types 1--4. \\
\midrule
\texttt{trafficking}     & Exploitation        & Arranging/facilitating travel for sexual exploitation. \\
\texttt{exploitation}    & Exploitation        & Sexual or labour exploitation via trafficking, coerced prostitution, or forced labour. \\
\texttt{prostitution\_related} & Exploitation  & Controlling prostitution for gain with compulsion, coercion, or force (\S54(2) SOA 2003). \\
\texttt{obscene\_publications} & Exploitation  & Distribution of obscene material (Obscene Publications Act 1959); includes extreme pornographic images. \\
\midrule
\texttt{other}           & Other               & Other forms of VAWG not covered above. \\
\bottomrule
\end{tabular}
\caption{VAWG crime-type taxonomy used for scenario generation, event graph
  annotation, and downstream classification. Definitions follow CPS VAWG
  strategy documentation.}
\label{tab:crime_taxonomy}
\end{table}
\twocolumn